\documentclass[runningheads]{llncs}
\usepackage[T1]{fontenc}
\usepackage{graphicx}
\usepackage{amsfonts}
\usepackage{amsmath}
\usepackage{hyperref}
\newcommand{\ptitle}[1]{{\bf #1}.}

\begin{document}
\title{How smoothing the affinity matrix affects neighborhood preservation in t-SNE}
\titlerunning{How smoothing affects neighborhood preservation in t-SNE}
%
\author{Shirin Mohebi\orcidID{0009-0008-1364-6173} \and
Guillaume Bied\orcidID{0009-0003-1370-711X} \and
Jefrey Lijffijt\orcidID{0000-0002-2930-5057}}
\authorrunning{S. Mohebi et al.}
%
\institute{Ghent University IDLab, Ghent, Belgium\\
\email{\{Shirin.Mohebi,Guillaume.Bied,Jefrey.Lijffijt\}@ugent.be}}
\maketitle              
\begin{abstract}
Dimensionality reduction methods are instrumental to visualize high-dimensional data, and t-SNE stands as one of the most widely used methods due to its emphasis on local neighborhood preservation. A central component of t-SNE is the \emph{affinity matrix}, which expresses pairwise similarities in the form of symmetrized probabilities, over which the optimization problem of t-SNE is defined. We study how the `sharpness' of this probability distribution affects neighborhood preservation at different scales. We introduce a row-wise power transform controlled by a parameter \(\gamma\) that can smooth or sharpen each row of the affinity matrix while preserving sparsity and rank order. We show that this transform is equivalent to rescaling the Gaussian bandwidth and thus to changing the perplexity. However, as the sharpness of the probability distribution varies per point, a fixed \(\gamma\) leads to point-dependent effective perplexities, making it distinct from changing the global perplexity. Empirically, we find that sharpening improves preservation of the very nearest neighbors, while smoothing improves preservation of broader local neighborhoods, outperforming alternative affinity constructions including multiscale methods in the mid-local range.
\end{abstract}
\section{Introduction}
\ptitle{Motivation} Machine learning models are highly effective for inference and prediction, but users often still want to inspect data directly to identify patterns and gain insights that complement model outputs. 
Visualization can support this kind of exploration, but this is difficult when the data are high-dimensional, meaning that each observation is described by many features. Dimensionality reduction (DR) methods mitigate this by projecting the data into a two dimensional (2D) space, after which we can visualize the data in a scatter plot.

\begin{figure}
    \centering
    \includegraphics[width=0.40\textwidth]{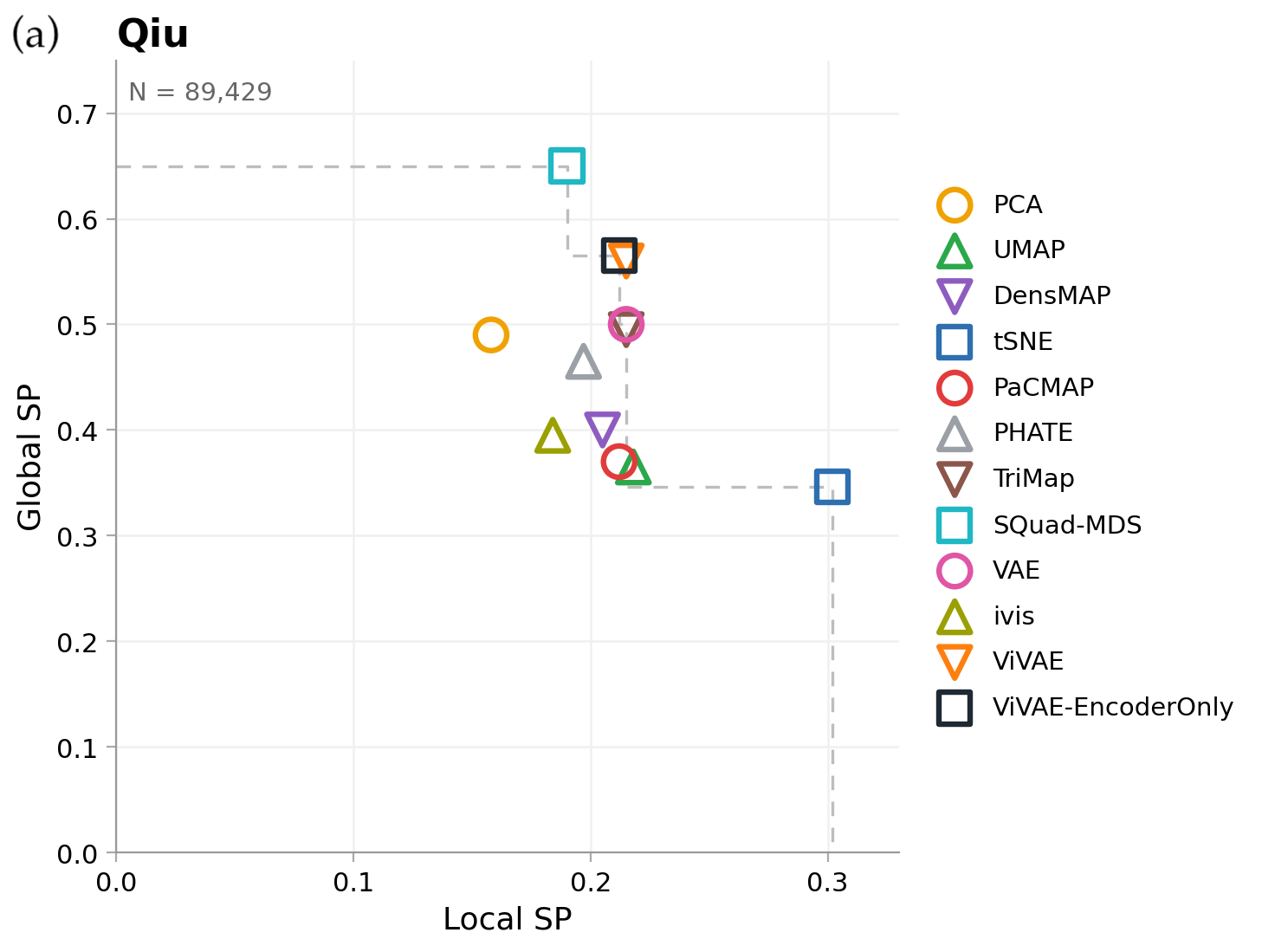} ~ 
    \includegraphics[width=0.40\textwidth]{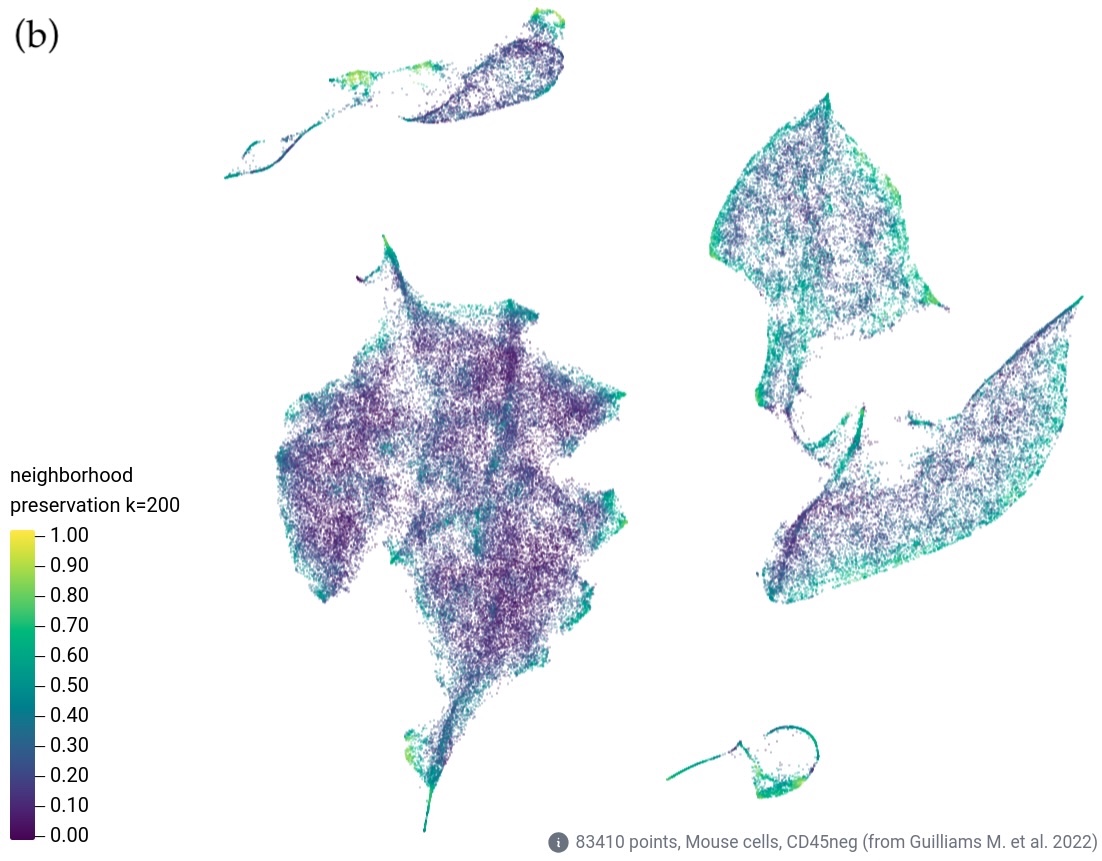}
    \caption{(a) Redrawn from Novak et al. (2026) \cite{Novak2026}, showing local versus global neighborhood preservation (here measured as the AUC under the RnX and log-RnX curve respectively, see their paper for details) for a range of state-of-the-art DR methods. One dataset shown here, the paper contains another seven. As can be seen, t-SNE manages to retain local structure substantially better than other DR methods. (b) From Heiter et al. (2024) \cite{heiter2024}, illustrating on a t-SNE projection of mouse cell data \cite{scheyltjens2022} that neighborhood preservation strongly varies across the embeddings. Color corresponds to neighborhood overlap with $k = 200$.\label{fig:SP}}
\end{figure}

Unfortunately, projecting data into 2D typically means that only part of the original structure is retained. A substantial amount of research has been conducted on developing effective DR methods for visualizing high-dimensional data. Among these methods, t-distributed Stochastic Neighbor Embedding (t-SNE) \cite{vandermaaten2008} is one of the most widely adopted nonlinear dimensionality reduction techniques \cite{debodt2025}. A major reason for its widespread use is its focus on local neighborhood structure: points that are close in the high-dimensional space are encouraged to remain close in the embedding. t-SNE is known to preserve local neighborhood structure well \cite{Novak2026}, see Figure~\ref{fig:SP}(a) for an illustration from the literature.
Although t-SNE often performs better than other methods on local preservation benchmarks, neighborhood preservation scores are often low in the absolute sense, and errors may also be spread unevenly over the embedding. See Figure~\ref{fig:SP}(b) for an illustration of this effect. It has been argued repeatedly that low reliability of (local) structure preservation creates problems in interpreting  apparent structures on the embedding, as they may be created artificially by the DR method, rather than derived from a reliable signal in the data \cite{chari2023}.

\ptitle{Aim of the study} As also shown in Figure~\ref{fig:SP}(a), other methods (e.g., PaCMAP \cite{Wang2021PaCMAP}, Autoencoders \cite{Novak2026}) outperform t-SNE on global structure preservation and most research appears to have focused on improving global structure preservation of t-SNE. However, it is equally interesting to ask \emph{whether it would be possible to improve local structure preservation further}, which may for example make the inspection of cluster substructures more reliable and useful. To the best of our knowledge, there has been no explicit investigation whether it is possible to improve neighborhood preservation of or beyond t-SNE. To this end, we study how t-SNE encodes local neighborhood relations in its high-dimensional affinity matrix and explore the impact of modifying the affinity scores.

\ptitle{Methodology} To better understand what guides t-SNE's local preservation behavior, we focus on its high-dimensional affinity matrix. t-SNE first converts high-dimensional distances into neighborhood probabilities, forming the affinity matrix \(P\). The embedding is then optimized so that the low-dimensional probabilities reproduce these high-dimensional affinities as closely as possible. After \(P\) has been constructed, the optimization is driven only by the affinities. This makes \(P\) a central component of t-SNE: it encodes which neighborhood relations are emphasized and how strongly they influence the embedding. Consequently, changes in the affinity matrix can change which local structures are preserved or lost. In this work, we investigate how changes in the smoothness of this affinity matrix affect t-SNE's neighborhood preservation.

Standard t-SNE uses point-specific Gaussian bandwidths to construct the affinity matrix, and these bandwidths are chosen so that every point has the same perplexity. Thus, t-SNE adapts the local distance scale while fixing the perplexity globally. The perplexity is \(2^{H(P_i)}\) with \(H(P_i)\) the Shannon entropy 
of the conditional probability row and generally interpreted as the effective number of neighbors \cite{vandermaaten2008}. However, we observe that typically only a few neighbors dominate the probability distribution for each point, i.e., that the distributions are `sharp', causing a few nearest neighbors to dominate the attractive forces. Moreover, the sharpness of the distributions varies per point. We influence this sharpness by applying a power transform controlled by a parameter \(\gamma\) that can smooth (or sharpen further) each row of the affinity matrix, while preserving the original neighbor support and rank order. This gives a controlled way of studying the effect of affinity sharpness on the final embedding.

\ptitle{Theoretical and empirical results} We show that the transform is equivalent to changing the effective Gaussian bandwidth, and hence the perplexity, but unlike increasing the global perplexity, it produces point-dependent effective 
perplexities. Empirically, we find that changing \(\gamma\) affects neighborhood preservation differently at different scales: sharpening improves preservation of the very nearest neighbors, while smoothing improves both preservation of broader local neighborhoods and global structure. The effect is most pronounced at lower perplexities, where the original affinity rows are sharpest, and the effect becomes smaller as perplexity increases. These improvements cannot be replicated by perplexity tuning alone, because smoothing assigns different effective neighborhood sizes to different points rather than just increasing perplexity of all points. We also show that \(\gamma\) smoothing outperforms alternative affinity constructions in the mid-local neighborhood range. Smoothing does not affect scalability as the number of (non-zero) neighbors in the affinity matrix is left unchanged.

\ptitle{Paper outline} We first discuss related work in Section~\ref{sec:relatedwork}. In Section~\ref{sec:methodology}, we revisit t-SNE, introduce the affinity smoothing operation controlled by $\gamma$, and characterize its effect on effective perplexity across data points. Section~\ref{sec:experiments} presents the empirical investigation of smoothing's effects on neighborhood preservation. Section~\ref{sec:conclusions} provides a summary of the main conclusions. Supplementary material, including an Appendix in PDF format, source figures, and replication code, is provided at \url{https://github.com/aida-ugent/smooth_affinity_tsne.git}.

\section{Related work\label{sec:relatedwork}}

t-distributed Stochastic Neighbor Embedding (t-SNE) \cite{vandermaaten2008} uses high-dimensional neighborhood probabilities from Gaussian kernels and then optimizes a low-dimensional embedding whose probabilities match those using a heavy-tailed Student \(t\) distribution. Kobak and Berens \cite{kobak2019} describe t-SNE as particularly effective in revealing local structure. Broader transcriptomic and cytometry benchmarks also report strong local-neighborhood preservation for t-SNE compared with methods such as UMAP \cite{McInnes2018UMAP}, TriMap \cite{Amid2019TriMap}, and PaCMAP \cite{Wang2021PaCMAP}. Similarly, Novak et al.~\cite{Novak2026} compare popular DR methods using separate local and global structure-preservation scores and find that t-SNE achieves the strongest local structure preservation among the evaluated methods. At the same time, these studies also show that local preservation is not perfect: even strong t-SNE embeddings can miss many high-dimensional neighbors \cite{chari2023}. This motivates analyses of which parts of the t-SNE pipeline control local-neighborhood preservation.

A large body of work exists on improving the practical use of t-SNE. Barnes-Hut t-SNE reduces the cost of the original algorithm using tree-based approximations \cite{vanderMaaten2013}, while FIt-SNE uses interpolation and fast Fourier transform techniques to further accelerate the computation \cite{linderman2019}. openTSNE provides a modular implementation that makes large-scale and experimental use of t-SNE more accessible \cite{Policar2024openTSNE}. Other work has studied optimization choices such as initialization, learning rate, early exaggeration, and iteration schedules \cite{Belkina2019OptTSNE,Chourasia2022InitKernelTSNE,kobak2019}. These works show that the quality and usability of t-SNE strongly depend on how the optimization is performed. Our focus is complementary to  these studies: we keep the optimization procedure fixed and study how the high-dimensional affinity distribution affects the embedding.

The work closest to ours concerns the construction of high-dimensional affinities and the choice of neighborhood scale. In standard t-SNE, the neighborhood scale is controlled by the target perplexity, making perplexity a central design choice in the affinity construction. Several works have questioned whether this single-scale construction is sufficient. Multiscale Stochastic Neighbor Embedding constructs multiscale similarities by averaging softmax-based similarities over multiple bandwidths, with the goal of improving embedding quality across scales and reducing dependence on a single scale parameter \cite{Lee2014MSSNE}. Perplexity-free t-SNE follows a related motivation by constructing neighborhoods across Gaussian kernels with growing widths, avoiding the need for a user-specified perplexity \cite{deBodt2018ttSNE}. Multi-scale kernels are also used in practical protocols for single-cell transcriptomics \cite{kobak2019}. Some implementations instead fix a single Gaussian bandwidth for all
points~\cite{Policar2024openTSNE}. Other variants modify the similarity kernels themselves. Twice Student t-SNE replaces the Gaussian high-dimensional similarities with Student-type ones, changing how neighborhoods are represented before optimization \cite{deBodt2018ttSNE}. Heavy-tailed kernel variants instead modify the low-dimensional Student-t kernel by changing its degrees of freedom, showing that heavier tails can reduce crowding and reveal finer cluster structure \cite{Kobak2019HeavyTailsTSNE}.

Our analysis is closest to this affinity-construction literature, but asks a narrower question: what is the effect of changing only the row-wise distribution of probability mass after the standard t-SNE affinities have been constructed?
We use a row-wise power transform as a controlled perturbation that preserves neighbor support and rank order while changing affinity sharpness. This allows us to isolate whether local neighborhood preservation depends only on which neighbors are present, or also on how strongly those neighbors are weighted.

\section{Methodology\label{sec:methodology}}
In this section we detail the proposed smoothing operation. The section is structured as follows. 
We first review t-SNE and the construction of the conditional probabilities that we later transform. We then introduce a power transform that can smooth (or sharpen) each conditional row, without changing which neighbors are present or how they are ordered. Finally, we show that this transform is equivalent to changing the bandwidth of the Gaussian distribution and introduce the concept of \emph{effective perplexity}, being the perplexity of the resulting distribution parameterized by the input perplexity \(\rho\) and the smoothing parameter \(\gamma\). This yields a generalized variant of t-SNE that enables us to study how the smoothness of the affinities affects neighborhood preservation in embeddings.

\subsection{Standard t-SNE affinity construction}

t-SNE is built up through three main concepts: (1) a mapping of the high-dimensional pairwise distances to (symmetrized) probabilities, collected in the affinity matrix, (2) definition of the similarity kernel in the low-dimensional space using the t-distribution, which is where the prefix \emph{t} comes from, and (3) the definition of the overall objective, namely the KL-divergence between the low- and high-dimensional probabilities. We review each in turn below.

\ptitle{From distances to conditional probabilities} The first step maps pairwise distances in the high-dimensional space to probabilities that represent similarities: Let $X = \{x_1, x_2, \ldots, x_n\}$ be the high-dimensional data, where \(x_i \in \mathbb{R}^m\). For each point \(x_i\), a conditional probability distribution is defined over the other points. The conditional probability of choosing \(x_j\) as a neighbor of \(x_i\) is
\begin{align}
p_{j|i} =
\frac{\exp(-\beta_i d_{ij}^2)}{\sum_{m \neq i} \exp(-\beta_i d_{im}^2)},
\qquad
\beta_i = \frac{1}{2\sigma_i^2}.
\label{eq:standard_conditional}
\end{align}
with \(p_{i|i}=0\) and  \(d_{ij} = \|x_i - x_j\|\), \(\sigma_i\) is the bandwidth of the Gaussian kernel centered at \(x_i\). A small \(\sigma_i\) produces a sharper distribution, while a larger \(\sigma_i\) produces a smoother distribution.

In t-SNE, a separate \(\sigma_i\) is identified for every point, so that each conditional distribution has a fixed user-defined perplexity: \(\mathrm{Perp}(P_i) = \rho\).
The perplexity of the conditional distribution \(P_i\) is defined as the exponential of its Shannon entropy where \(P_i = \{p_{j|i}\}_{j \neq i}\) denotes row \(i\) of the conditional probability (affinity) matrix and it contains the neighborhood relations for point $i$:
\begin{align}
\mathrm{Perp}(P_i)
=
2^{H(P_i)},
\qquad
H(P_i)
=
-\sum_j p_{j|i}\log_2 p_{j|i}.
\label{eq:perplexity}
\end{align}

\(\mathrm{Perp}(P_i)\) is commonly interpreted as the effective number of neighbors represented by the distribution \cite{vandermaaten2008}, and equals
\(\rho\) for all points.

\ptitle{The affinity matrix} After computing the conditional probabilities, t-SNE symmetrizes them to obtain a joint high-dimensional \emph{affinity matrix} $P$ by setting $ p_{ij}=
(p_{j|i}+p_{i|j})/(2n)$ for $i \neq j$, and $p_{ii}=0$. The entries \(p_{ij}\) define the strength of the high-dimensional neighborhood relation between points \(i\) and \(j\). Larger values of \(p_{ij}\) indicate stronger attractive relations in the embedding optimization.

\ptitle{Similarity in the embedding}
t-SNE then defines a corresponding probability distribution in the low-dimensional embedding. Let
\(Y =\) \(\{y_1, y_2, \ldots, y_n\}\), where \(y_i \in \mathbb{R}^2\). The low-dimensional similarity between two embedded points is defined using a heavy-tailed Student \(t\) kernel with one degree of freedom:
\begin{align}
q_{ij}
=
\frac{
\left(1+\|y_i-y_j\|^2\right)^{-1}
}{
\sum_{k \neq l}
\left(1+\|y_k-y_l\|^2\right)^{-1}
},
\qquad q_{ii}=0.
\label{eq:low_dim_affinity}
\end{align}
The heavy-tailed kernel allows moderately distant points in the embedding to still have non-negligible similarity, which helps reduce the crowding problem.

\ptitle{The optimization objective}
Finally, the embedding \(Y\) is found by minimizing the Kullback--Leibler divergence between the high-dimensional distribution \(P\) and the low-dimensional distribution \(Q\):
\begin{align}
C
=
KL(P \| Q)
=
\sum_{i \neq j}
p_{ij}
\log
\frac{p_{ij}}{q_{ij}}.
\label{eq:kl_objective}
\end{align}
This objective encourages pairs with large \(p_{ij}\) to also have large \(q_{ij}\), meaning that points with strong high-dimensional affinities should remain close in the embedding. Therefore, the high-dimensional affinity matrix \(P\) determines which local neighbor relations are emphasized during optimization.

\subsection{Row-wise smoothing of conditional probabilities}
Standard t-SNE fixes the target perplexity of every conditional probability row. However, this does not necessarily mean that the resulting affinity rows assign probability mass evenly across neighbors.

\begin{figure}[t]
    \centering
    \begin{minipage}{0.48\linewidth}
        \includegraphics[width=\linewidth]{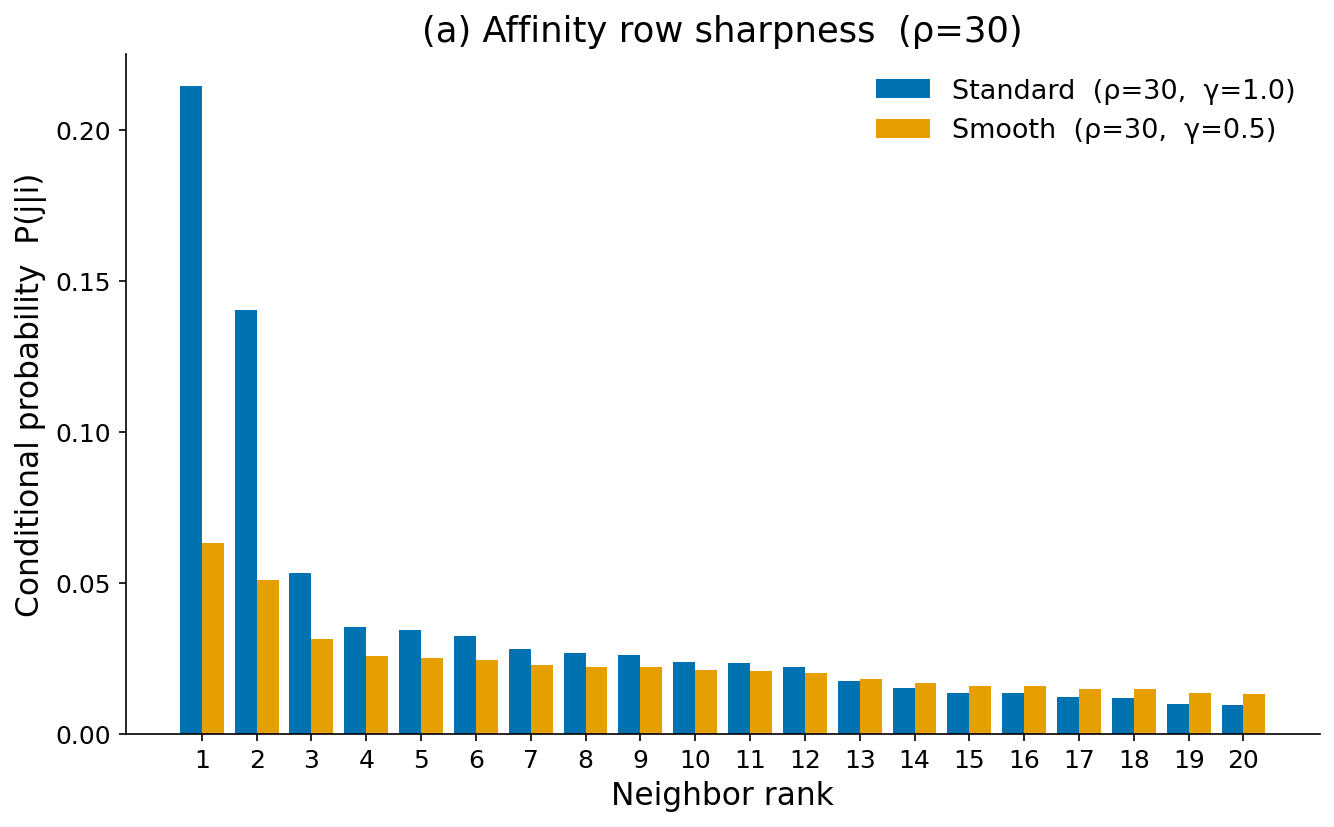}
        \label{fig:single_row}
    \end{minipage}
    \hfill
    \begin{minipage}{0.48\linewidth}
        \includegraphics[width=\linewidth]{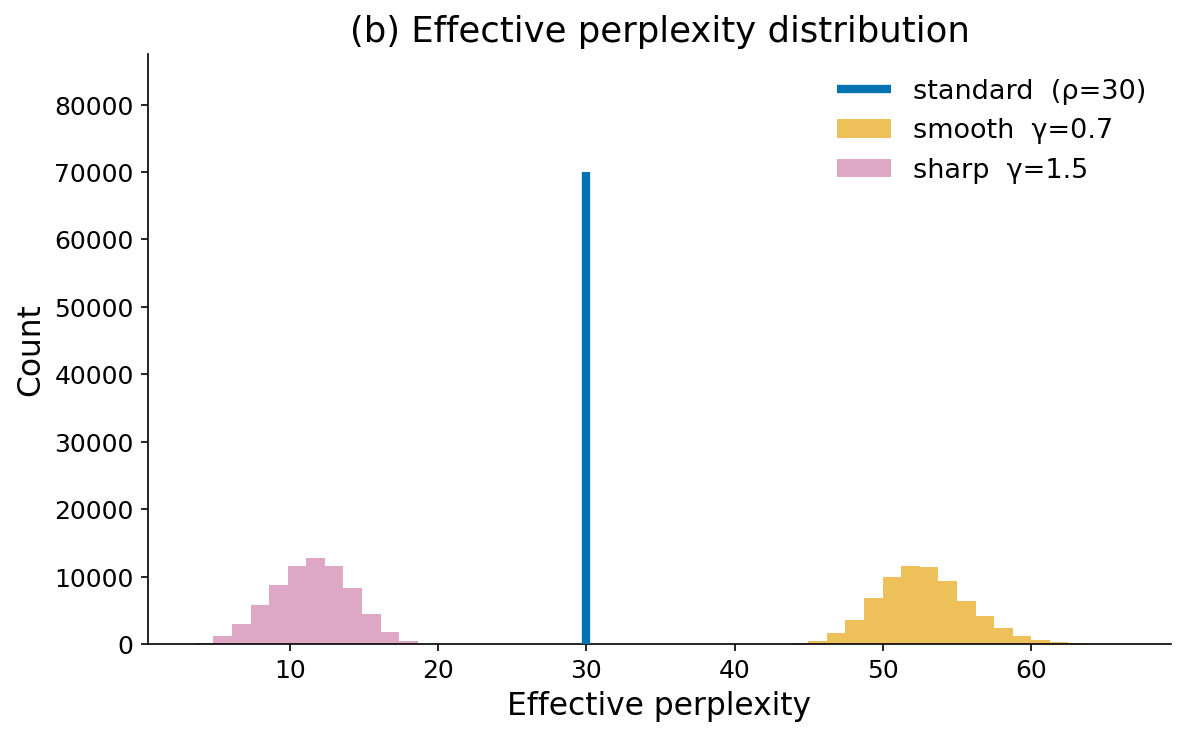}
        \label{fig:cumulative_mass}
    \end{minipage}
    \caption{(a) Affinity row sharpness at \(\rho=30\) for a single point from MNIST (selected as average wrt.\ summed mass of the top-5 neighbors). Standard t-SNE (\(\gamma=1.0\)) drops steeply after the first few neighbors, while smoothing (\(\gamma=0.5\)) redistributes probability mass more evenly across the neighborhood. (b) Distribution of Effective perplexity across points on MNIST for standard t-SNE (\(\rho=30\)), smoothed t-SNE (\(\gamma=0.7\)), and sharpened t-SNE (\(\gamma=1.5\))\label{fig:affinity_illustration}}
\end{figure}

From early empirical analysis, we observed that the affinity distributions are typically very skewed: as illustrated in Figure~\ref{fig:affinity_illustration}(a), at the default perplexity of 30 just a few neighbors dominate. This led us to the idea of smoothing the distribution to spread probability mass more equally over the neighbors.

We apply the following power transform to the conditional probabilities:
\begin{align}
\tilde p_{j|i}
=
\frac{p_{j|i}^{\gamma}}
{\sum_{m \neq i} p_{m|i}^{\gamma}},
\label{eq:power_smoothing}
\end{align}
where \(\gamma \geq 0\) controls the sharpness of the row distribution.
When \(0 \le \gamma < 1\), the transform smooths the distribution: large probabilities are reduced relative to smaller probabilities, and probability mass is redistributed toward lower-probability neighbors. When \(\gamma > 1\), the transform sharpens the distribution by concentrating more mass on the largest probabilities. When \(\gamma = 1\), the original t-SNE conditional probabilities are recovered.

As shown in Figure~\ref{fig:affinity_illustration}(a), applying the power transform with \(\gamma=0.5\) redistributes the mass more evenly and all neighbors now receive meaningful probability mass.
Note that it is common to use \(3\cdot\rho\) neighbors in the optimization and define all further neighbors to have \(p_{j|i} = 0\)~\cite{vanderMaaten2013,Policar2024openTSNE}. We do not make any change with respect to this, and hence runtime will be largely unaffected by smoothing.

This transformation preserves the rank order within each row, because the power transform is monotonic for \(\gamma > 0\) and therefore changes only the relative probability mass.
After applying the row-wise transform, the modified conditional probabilities are symmetrized in the same way as in standard t-SNE, namely by setting $\tilde p_{ij} =
(\tilde p_{j|i}+\tilde p_{i|j})/(2n)$ when $i \neq j$, and $\tilde p_{ii}=0$.
The resulting matrix \(\tilde P\) is then used in the standard t-SNE objective: $\tilde C
=
KL(\tilde P \| Q)$.
We leave the low-dimensional kernel and optimization objective unchanged.

\subsection{Bandwidth and effective-perplexity interpretation}
\ptitle{Smoothing changes the effective bandwidth} The smoothing transform in Eq.~\ref{eq:power_smoothing} has a direct interpretation in terms of the Gaussian bandwidths used in the high-dimensional affinity construction. Starting from the standard t-SNE conditional probabilities in Eq.~\ref{eq:standard_conditional}, applying the row-wise power transform gives
\begin{align*}
\tilde p_{j|i}
=
\frac{
\left(
\frac{\exp(-\beta_i d_{ij}^2)}
{\sum_{r \neq i} \exp(-\beta_i d_{ir}^2)}
\right)^{\gamma}
}{
\sum_{m \neq i}
\left(
\frac{\exp(-\beta_i d_{im}^2)}
{\sum_{r \neq i} \exp(-\beta_i d_{ir}^2)}
\right)^{\gamma}
}.
\end{align*}
The denominator of the original conditional distribution, $Z_i=\sum_{r \neq i} \exp(-\beta_i d_{ir}^2)$, is constant within row \(i\). Therefore, the factor \(Z_i^{-\gamma}\) appears in both the numerator and the denominator of the transformed row and cancels out during renormalization. This gives
\[
\tilde p_{j|i}
=
\frac{
\exp(-\gamma \beta_i d_{ij}^2)
}{
\sum_{m \neq i} \exp(-\gamma \beta_i d_{im}^2)
}.
\]
Thus, applying the power transform to the conditional probabilities is equivalent to replacing the original Gaussian precision \(\beta_i\) by $\tilde \beta_i = \gamma \beta_i$. Since \(\beta_i = 1/(2\sigma_i^2)\), we can write
$\tilde \beta_i
=
\frac{1}{2\tilde\sigma_i^2}
=
\gamma \frac{1}{2\sigma_i^2}$. Solving for the new effective bandwidth gives

\begin{equation}
\tilde\sigma_i
=
\frac{\sigma_i}{\sqrt{\gamma}}.
\label{eq:sigma_transform}
\end{equation}

Therefore, when \(0 \le \gamma < 1\), the effective bandwidth increases, i.e., \(\tilde\sigma_i > \sigma_i\). This means that smoothing makes the effective Gaussian kernel wider and the resulting conditional distribution less concentrated. Conversely, when \(\gamma > 1\), the effective bandwidth decreases, making the distribution sharper.

\ptitle{Smoothing yields point-dependent effective perplexities}
In standard t-SNE, the bandwidths \(\sigma_i\) are chosen so that each conditional probability row \(P_i=\{p_{j|i}\}_{j\neq i}\) matches the same target perplexity \(\rho\). After applying the transform in Eq.~\ref{eq:power_smoothing}, the \emph{effective perplexity} of the smoothed row \(\tilde P_i=\{\tilde p_{j|i}\}_{j\neq i}\) is
\begin{align*}
\mathrm{Perp}(\tilde P_i)
=
2^{H(\tilde P_i)},
\qquad
H(\tilde P_i)
=
-\sum_j \tilde p_{j|i}\log_2 \tilde p_{j|i}.
\end{align*}
Substituting Eq.~\ref{eq:power_smoothing} into the entropy of the smoothed row gives
\[
H(\tilde P_i)
=
-\sum_j
\frac{p_{j|i}^{\gamma}}
{\sum_{m\neq i}p_{m|i}^{\gamma}}
\log_2
\left(
\frac{p_{j|i}^{\gamma}}
{\sum_{m\neq i}p_{m|i}^{\gamma}}
\right).
\]

This expression depends on the individual probabilities \(p_{j|i}\), not only on the original entropy \(H(P_i)\) that is fixed by the constant perplexity $\rho$. Applying a constant \(\gamma \neq 1\) across rows can therefore produce different smoothed entropies, and hence different effective perplexities.

This gives the adaptive-perplexity interpretation of the transform: row-wise smoothing lets perplexity differ across rows, unlike standard t-SNE. Smoothing is thus also a different operation than running standard t-SNE with a larger perplexity, which would again match every row to the same new target perplexity.

Figure~\ref{fig:affinity_illustration}(b) confirms this empirically on MNIST. It shows the distribution of effective perplexities across points for standard t-SNE (\(\gamma=1\)) and smoothed and sharpened distributions, all starting from perplexity \(\rho=30\). Standard t-SNE concentrates all points at exactly the target perplexity of \(30\). In contrast, smoothing with \(\gamma=0.7\) produces a broad distribution of effective perplexities centered around \(53\), and sharpening with \(\gamma=1.5\) produces a broad distribution centered around \(12\). 

Since the change in effective perplexity depends on the original row distribution, it naturally varies across points: points with sharper affinity rows receive a larger increase, while points in sparser regions tend to be more affected as well (see online Appendix Fig. 2 for details). Together, these results confirm that \(\gamma\) controls a point-adaptive redistribution of probability mass, producing a heterogeneous spread of effective neighborhood sizes rather than a uniform shift.

\section{Experiments\label{sec:experiments}}

We designed the experiments to address the following research questions:
\begin{enumerate}
\item Does smoothing affect neighborhood preservation, and at what neighborhood scale? (Section~\ref{sec:effect})
\item Is it possible to achieve the same effect by changing the global perplexity? (Section~\ref{sec:comparison})
\item How do \(\gamma\) and $\rho$ affect neighborhood preservation of near-local and mid-local neighborhood scales? (Section~\ref{sec:sensitivity})
\item Does smoothing also improve global structure preservation? (Section~\ref{sec:global})
\item How does \(\gamma\) smoothing compare to alternative affinity 
constructions in terms of neighborhood preservation? 
(Section~\ref{sec:affinity_comparison})
\end{enumerate}

We used three datasets: MNIST~\cite{lecun1998mnist} (\(n=70{,}000\), \(m=784\)), UCI Adult~\cite{kohavi1996adult} (\(n=48{,}842\), \(m=14\)), and the mouse cortex dataset from Tasic et al.~\cite{Tasic2018} (\(n=23{,}822\), \(m=50\)). For MNIST, pixel values are scaled to \([0,1]\) and reduced to 50 principal components, following Kobak and Berens~\cite{kobak2019}. For Adult, numerical variables are standardized and categorical variables are one-hot encoded. For mouse cortex, we use the preprocessed data from Kobak and Berens~\cite{kobak2019}.
Due to space constraints, we show results for MNIST only. Full results are available in the online appendix and are consistent across the three datasets.

All embeddings are computed using openTSNE~\cite{Policar2024openTSNE}, and we implemented \(\gamma\) as a parameter of the affinity construction within the forked version at \url{https://github.com/aida-ugent/smooth_affinity_tsne.git}. For fair comparison, all variants use the same initialization, early exaggeration, learning rate, number of iterations, and optimization settings. The only difference between standard t-SNE and the smoothed variants is the row-wise transformation.
All reported metric values are averages over five runs. Standard deviations are omitted as too small to visualize: below \(0.003\) (typically \(0.0005\)) for every reported \(NO@k\) and AUC across all three datasets. For global structure preservation they are larger, so bands are shown in
Figure~\ref{fig:global_nh_affinity}(a). All results and per-run values are in
the online repository.

\subsection{Effect of smoothing on local neighborhood preservation}
\label{sec:effect}
To measure neighborhood structure preservation, we use neighborhood overlap at \(k\), denoted \(NO@k\) (also known as \(Q_{NX}(k)\) in the DR evaluation literature~\cite{Lee2009QNX}), which measures the fraction of high-dimensional \(k\)-nearest neighbors that are also among the \(k\)-nearest neighbors in the embedding.
\[
NO_i(k)
=
\frac{
|\mathcal{N}^{HD}_i(k) \cap \mathcal{N}^{LD}_i(k)|
}{k},
\qquad
NO@k
=
\frac{1}{n}
\sum_{i=1}^{n}
NO_i(k).
\]

\begin{figure}[tp]
    \centering
    \includegraphics[width=0.6\linewidth]{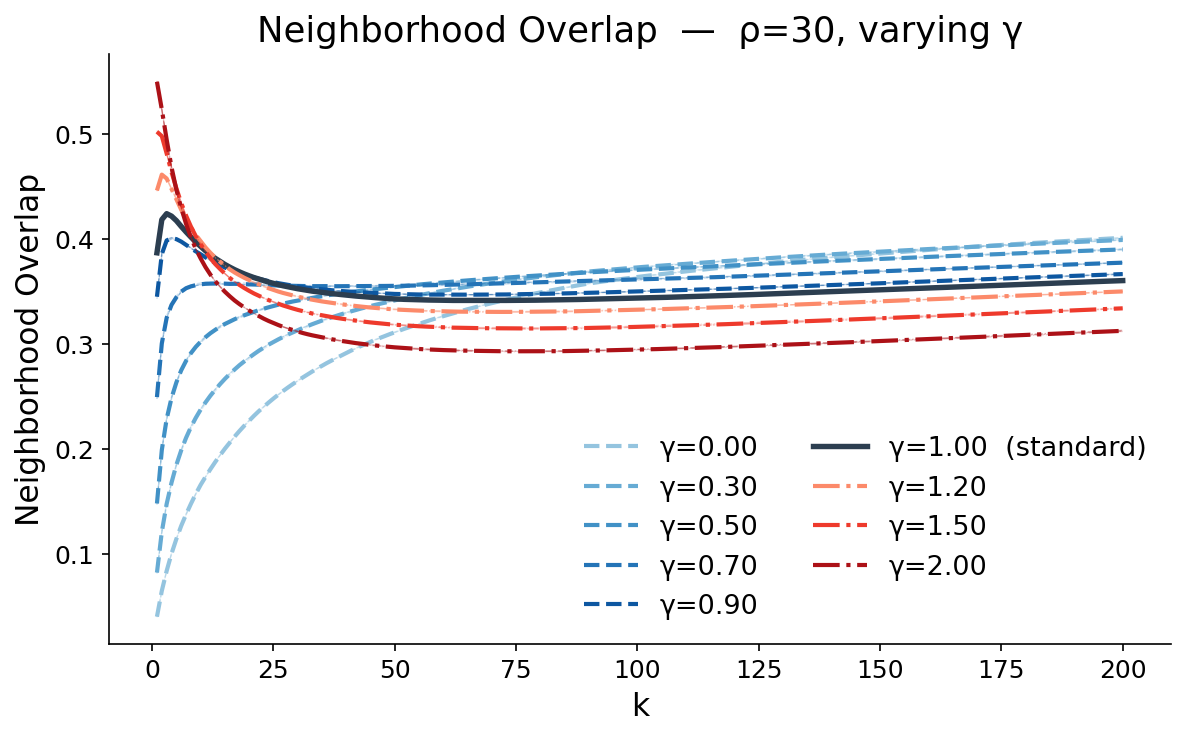}
    \caption{\(NO@k\) for \(k=1,\ldots,200\) at perplexity \(\rho=30\) 
    with varying \(\gamma\), on MNIST. Sharpened variants (\(\gamma > 1\)) 
    perform better for small \(k\), while smoothed variants (\(\gamma < 1\)) perform
    better for larger \(k\) (starting from $k \approx 2/3 \rho$).}
    \label{fig:main_mnist}
\end{figure}

Figure~\ref{fig:main_mnist} shows \(NO@k\) for \(k=1,\ldots,200\) at perplexity \(\rho=30\) for MNIST. Smoothed variants (\(\gamma < 1\)) score better at higher values of \(k\), with the gap growing as \(k\) increases, while sharpened variants (\(\gamma > 1\)) score better at small \(k\). 

\begin{figure}[tp]
    \centering
    \includegraphics[width=0.8\linewidth]{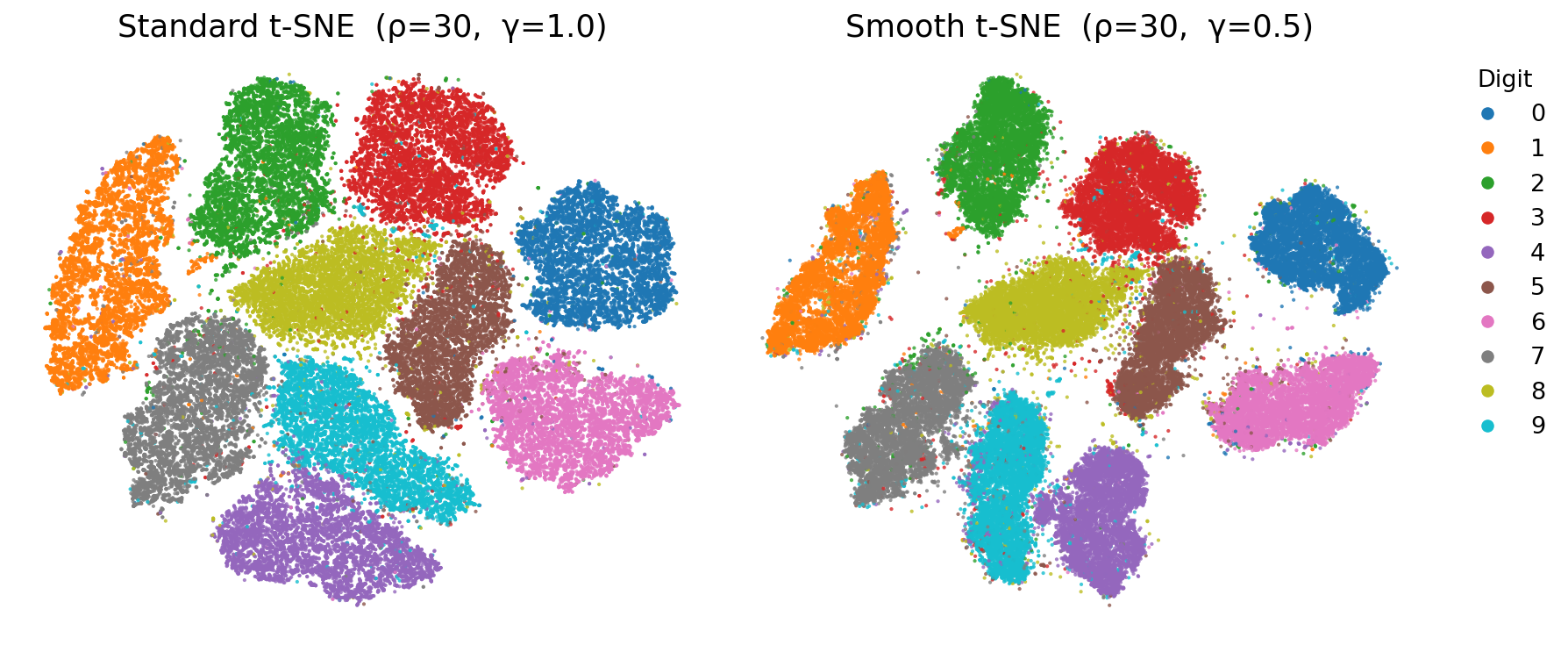}
    \caption{
    Standard t-SNE (\(\rho=30\)) and smoothed t-SNE (\(\rho=30\), 
    \(\gamma=0.5\)) embeddings of MNIST. Smoothing produces clearer 
    inter-cluster separation while preserving the overall digit cluster structure. 
    }
    \label{fig:embedding}
\end{figure}

Figure~\ref{fig:embedding} shows the embeddings produced by standard t-SNE and smoothed t-SNE (\(\gamma=0.5\)) on MNIST. The overall cluster structure is preserved in both cases, but the smooth version shows stronger inter-cluster separation: the dense overlapping region in the center is broken up, digit 0 (blue) becomes clearly isolated, and several neighboring digit classes are pulled apart. This is consistent with smoothing giving more weight to broader neighborhoods, which pushes clusters further apart in the embedding.

\begin{figure}[tp]
    \centering
    \includegraphics[width=0.5\linewidth]{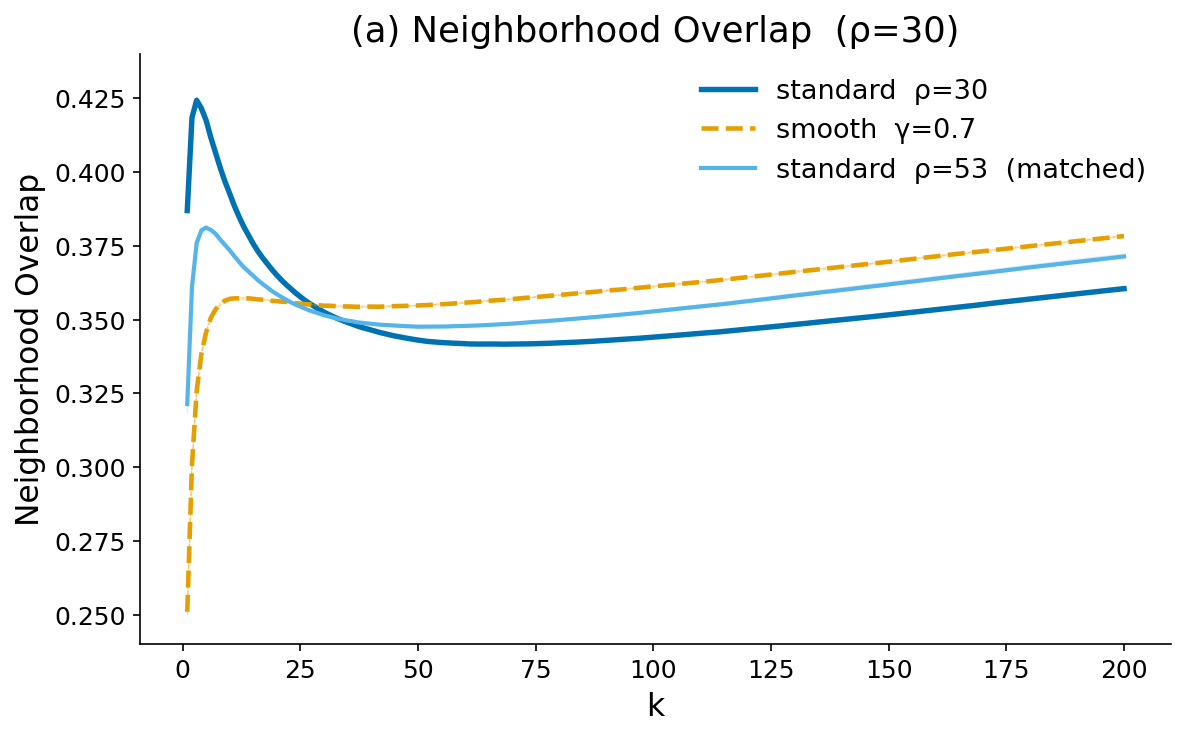}
    \hfill
    \includegraphics[width=0.44\linewidth]{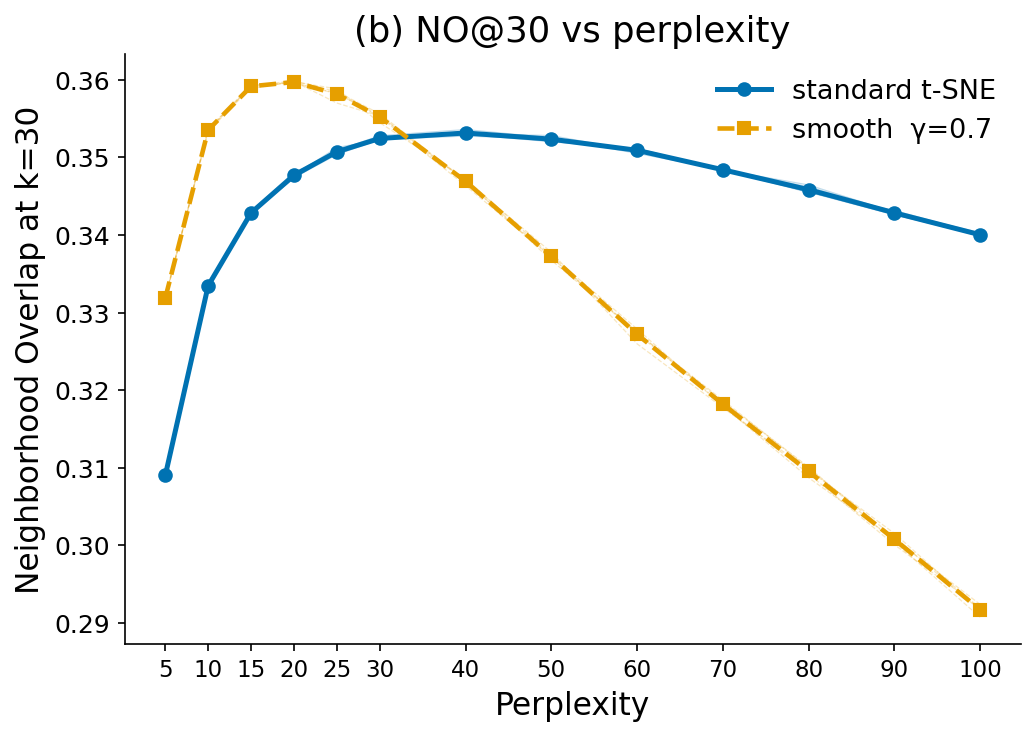}
    \caption{(a) NO@k comparison of standard t-SNE (\(\rho=30\)), 
    smoothed t-SNE (\(\rho=30\), \(\gamma=0.7\)), and standard t-SNE at \(\rho=53\), matching the median effective perplexity of smoothed t-SNE.
    (b) $NO@30$ for standard and smoothed t-SNE (with constant $\gamma=0.7$), for varying values of $\rho$. We observe that both standard t-SNE with same perplexity and matched perplexity have lower broad-local NO. Comparing across perplexities for a specific $NO@k$, we see that standard t-SNE with higher perplexities also does not achieve the same improved NO.
    \label{fig:comparison}}
\end{figure}

\subsection{Smoothing versus increasing perplexity\label{sec:comparison}}

A natural question is whether the increased neighborhood preservation at larger $k$ can also be achieved by increasing the perplexity. We showed in Section~\ref{sec:methodology} that smoothing produces point-dependent effective perplexities, but the question is whether that matters for neighborhood preservation.

To this end, Figure~\ref{fig:comparison}(a) shows three variants: standard t-SNE at \(\rho=30\), smoothed t-SNE at \(\rho=30\) with \(\gamma=0.7\), and standard t-SNE at \(\rho=53\), chosen to match the median effective perplexity across points of the smoothed variant (see Appendix~D). For \(k > 25\), smoothed t-SNE consistently outperforms standard t-SNE at \(\rho=30\) and \(\rho=53\). This confirms that the point-dependent effective perplexities produced by smoothing lead to different and better mid-local preservation than simply increasing the global perplexity.

However, we are unsure which perplexity would be best for which $NO@k$. To explore this through an example, Figure~\ref{fig:comparison}(b) shows $NO@30$, while varying perplexity for standard and smoothed t-SNE (with constant $\gamma = 0.7$; not tuned). We observe that smoothing strongly affects which perplexity is optimal for this specific value of $k$. For standard t-SNE the best perplexity is around 40, while for smoothed t-SNE with $\gamma=0.7$ the best value is achieved with perplexity 20. We also confirm that a higher value of $NO@30$ is achieved for smoothed t-SNE (\(0.360\)) than standard t-SNE (\(0.353\)), confirming that smoothed t-SNE cannot be emulated by simply changing the perplexity.

\subsection{Sensitivity to \(\gamma\) and \(\rho\)}
\label{sec:sensitivity}

This experiment studies the results' sensitivity to smoothing  parameter \(\gamma\) and perplexity \(\rho\). We consider $\rho \in \{30, 50, 100, 200\}$ and $\gamma \in \{0, 0.5, 0.7, 1, 1.2, 1.5, 2.0\}$. The extreme case \(\gamma=0\) produces a uniform distribution over all neighbors, i.e. discards the rank order within each row.

\begin{figure}[t]
    \centering
    \begin{minipage}{0.48\linewidth}
        \includegraphics[width=\linewidth]{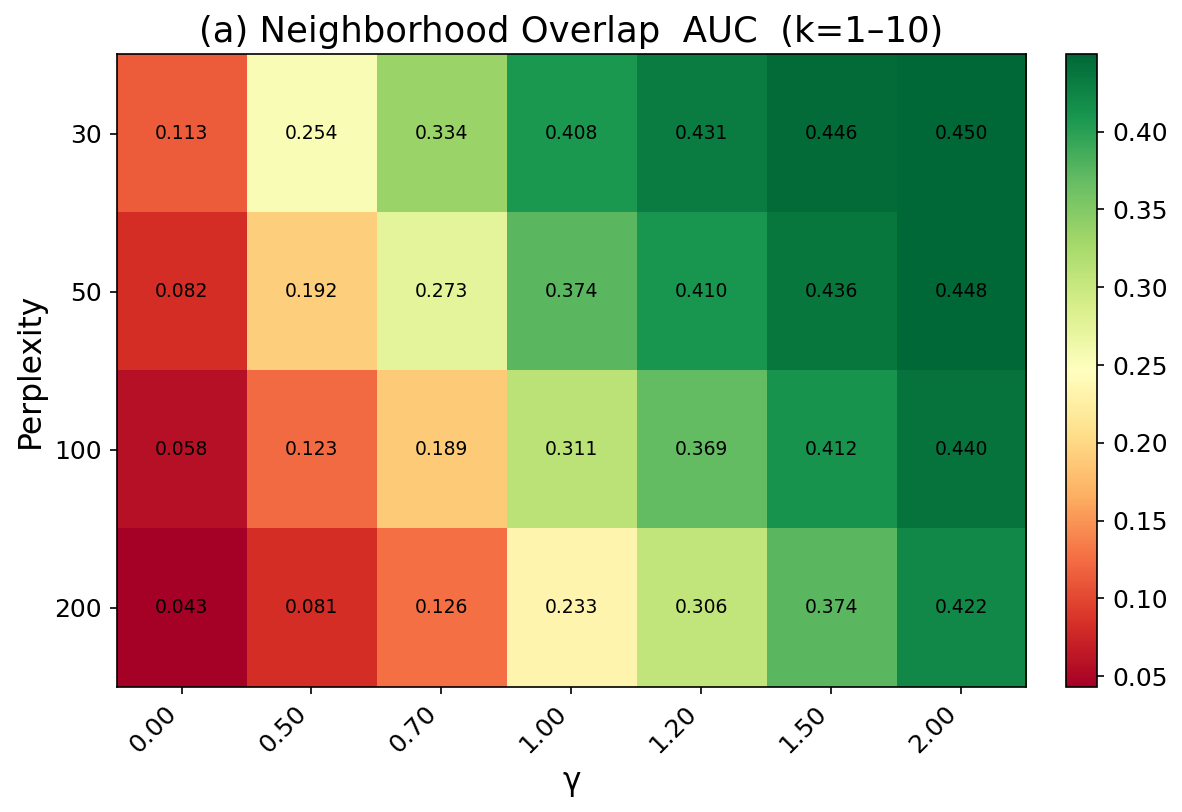}
        \label{fig:sensitivity_a}
    \end{minipage}
    \hfill
    \begin{minipage}{0.48\linewidth}
        \includegraphics[width=\linewidth]{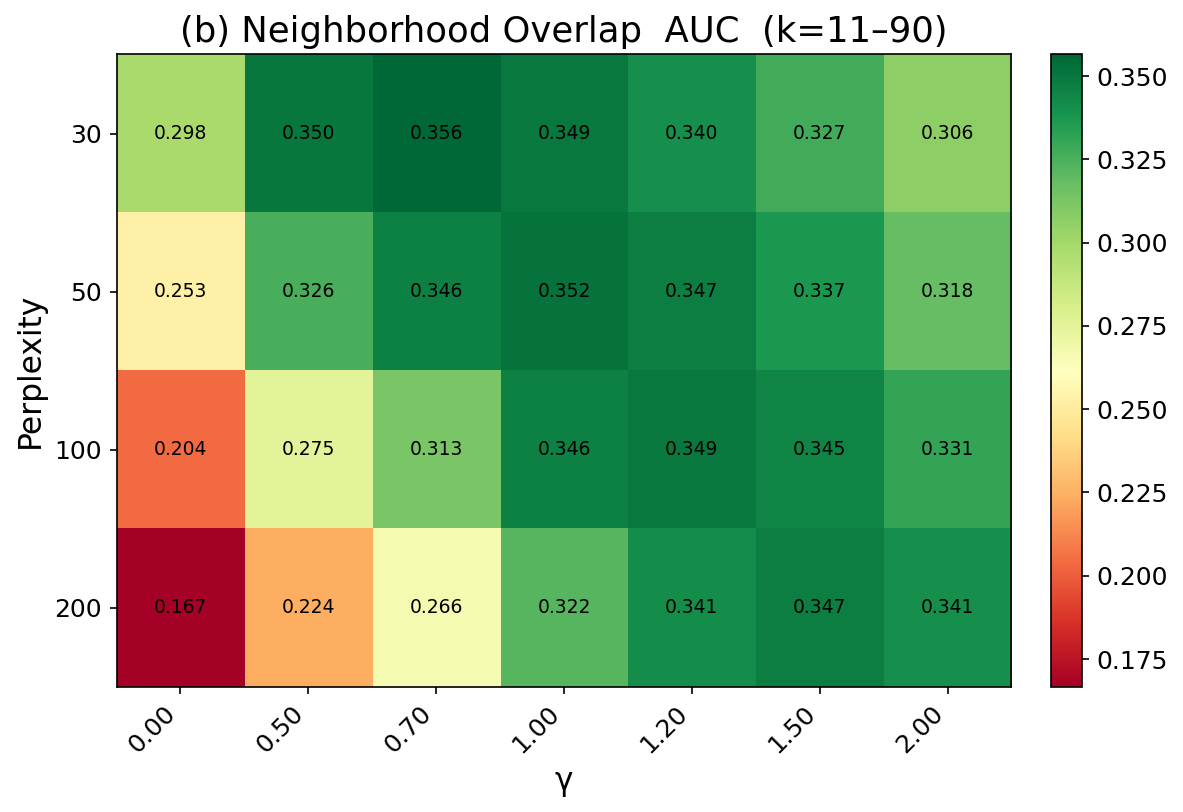}
        \label{fig:sensitivity_b}
    \end{minipage}
    \caption{Sensitivity of t-SNE behavior to \(\gamma\) and \(\rho\) on the MNIST dataset. Each cell shows the metric value 
    for a combination of \(\rho\) (rows) and \(\gamma\) 
    (columns). (a) Near-local preservation improves consistently with 
    both sharpening and lowering the perplexity. (b) Mid-local preservation peaks at moderate smoothing around \(\gamma=0.7\); extreme smoothing with \(\gamma=0\) performs worst.}
    \label{fig:sensitivity}
\end{figure}

We examined the effect of these parameters on near-local structure (\(k=1,\ldots,10\)) and mid-local structure (\(k=11,\ldots,90\); arbitrary cut-offs), shown in Figure~\ref{fig:sensitivity}. We summarize neighborhood
preservation for each range using the area under the \(NO@k\) curve (AUC). For near-local preservation (Figure~\ref{fig:sensitivity}a), sharpening consistently improves performance across all perplexity settings. The best value of AUC is \(0.450\) at \(\rho=30\), \(\gamma=2\), while the worst is \(0.043\) at \(\rho=200\), \(\gamma=0\). Higher \(\rho\) also hurts near-local preservation, regardless of \(\gamma\).

For mid-local preservation (Figure~\ref{fig:sensitivity}b), the pattern depends on the initial perplexity. At lower perplexities, moderate smoothing outperforms standard t-SNE: the best value is \(0.356\) at \(\rho=30\), \(\gamma=0.7\) (highest overall). However, at higher perplexity, the optimal \(\gamma\) shifts toward standard or even slightly sharpened variants. At \(\rho=200\), the best value is \(0.347\) at \(\gamma=1.5\). As expected, \(\gamma=0\) performs worst across all perplexity 
settings, reaching only \(0.167\) at \(\rho=200\).

\begin{figure}[t]
    \centering
    \begin{minipage}{0.48\linewidth}
        \includegraphics[width=\linewidth]{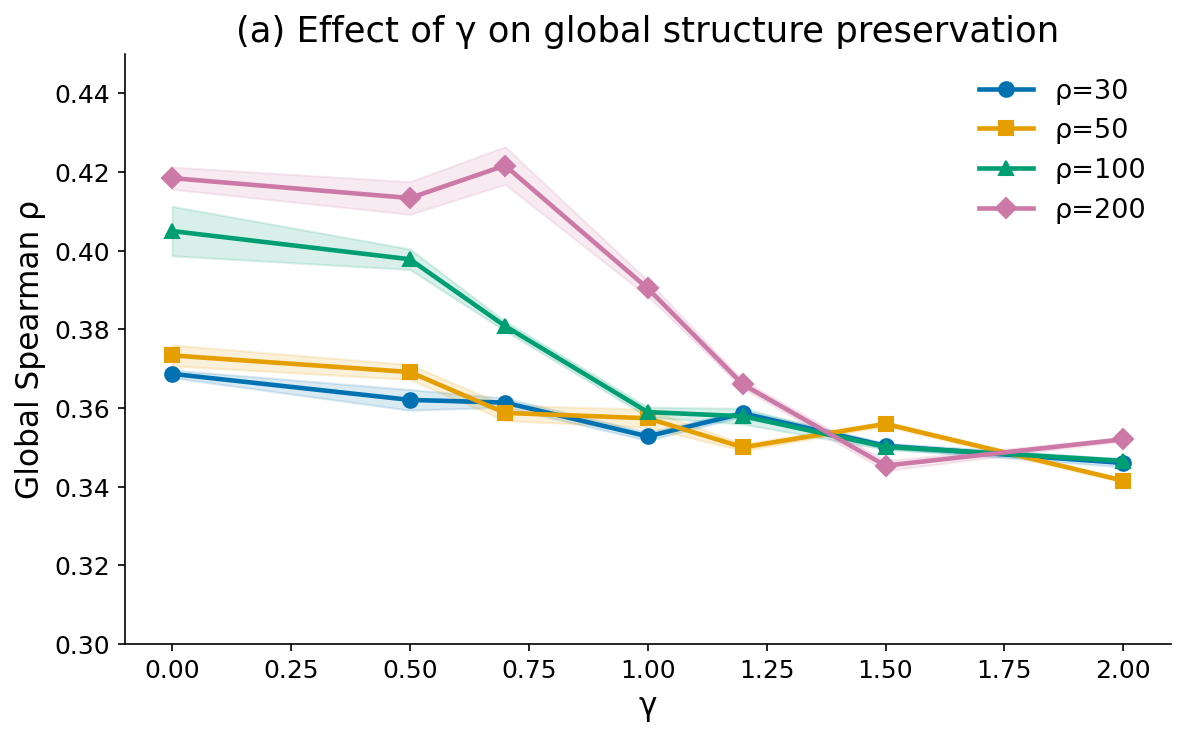}
        \label{fig:global_spearman}
    \end{minipage}
    \hfill
    \begin{minipage}{0.48\linewidth}
        \includegraphics[width=\linewidth]{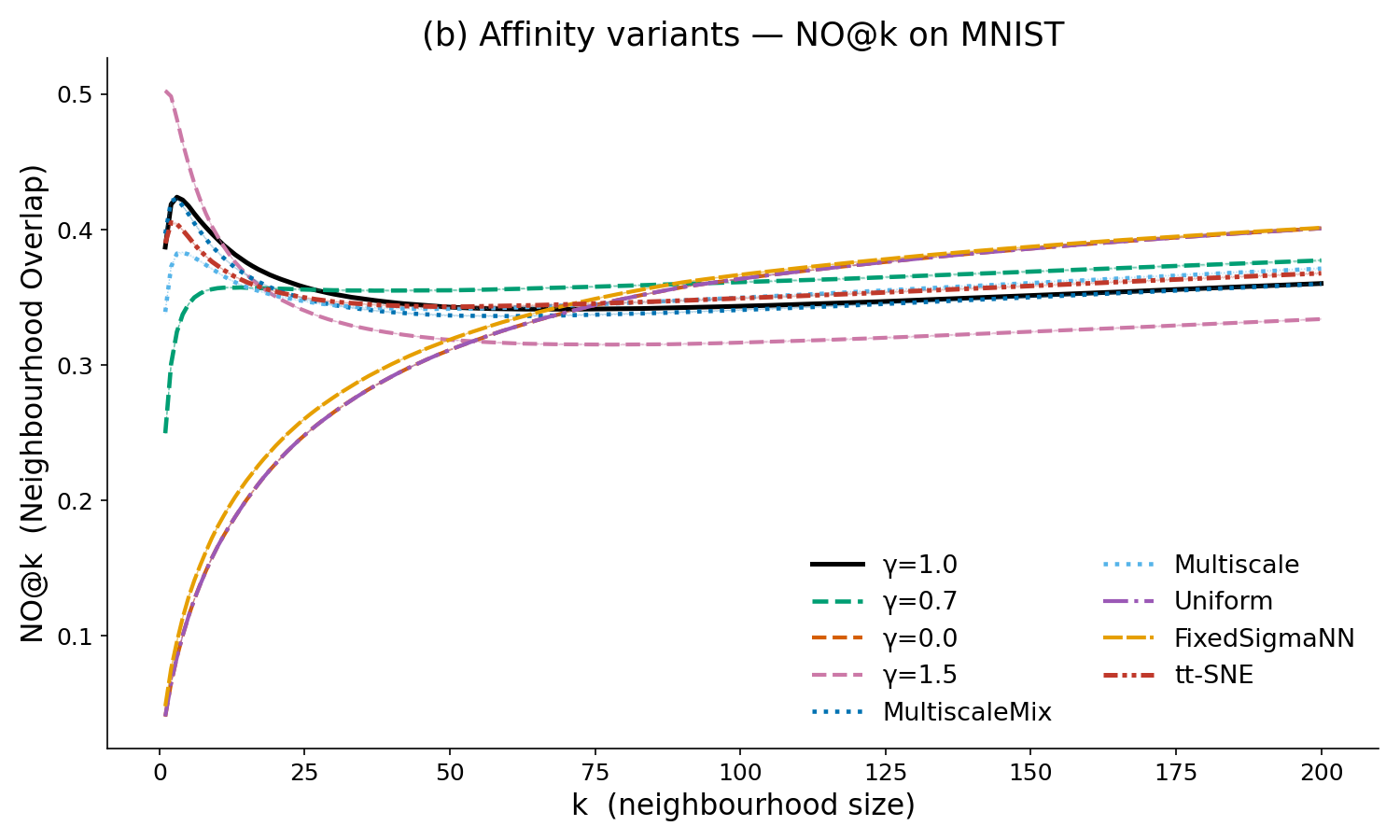}
        \label{fig:affinity_comparison}
    \end{minipage}
    \caption{ (a) Effect of \(\gamma\) on global structure preservation 
    (Spearman \(\rho\) between high- and low-dimensional distances) across four perplexity values on MNIST. In general, lower \(\gamma\) and higher \(\rho\) improve global structure preservation. Shaded bands indicate ±1 SD across reruns. (b) \(NO@k\) for \(k=1,\ldots,200\) on MNIST, comparing \(\gamma\) variants against alternative affinity constructions. Sharpening (\(\gamma=1.5\)) leads at small \(k\); smoothing (\(\gamma=0.7\)) outperforms all methods in the mid-local range; \(\gamma=0.0\)/\texttt{Uniform} and \texttt{FixedSigmaNN} dominate at large \(k\).}
    \label{fig:global_nh_affinity}
\end{figure}

\subsection{Effect of smoothing on global structure preservation}
\label{sec:global}

Figure~\ref{fig:global_nh_affinity} (a) shows the effect of \(\gamma\) on global structure preservation, measured through the Spearman rank correlation between high- and low-dimensional distances~\cite{Joia2011LAMP}. Generally, smoothing improves global structure preservation while sharpening degrades it. As expected, larger perplexities also lead to stronger global structure preservation, but the effect of smoothing is also stronger at higher perplexities. The best result is achieved at \(\rho=200\), \(\gamma=0.75\).

\subsection{Comparison with alternative affinity constructions}
\label{sec:affinity_comparison}

We compare affinity smoothing against five affinity constructions 
available in openTSNE~\cite{Policar2024openTSNE}.
\texttt{PerplexityBasedNN}~\cite{vandermaaten2008} 
is the standard t-SNE affinity, equivalent to our \(\gamma=1.0\). 
\texttt{MultiscaleMixture} and \texttt{Multiscale}~\cite{kobak2019} both 
incorporate multiple bandwidth scales: the former uses a mixture of Gaussians 
at different perplexities, while the latter averages single-scale probability 
distributions across scales. \texttt{Uniform} assigns equal probability mass 
to all \(k\) nearest neighbors, which should be equivalent to \(\gamma=0\). 
\texttt{FixedSigmaNN} uses a fixed Gaussian bandwidth across all points rather 
than a perplexity-adaptive one, which also results in point-dependent effective 
perplexities. We additionally include tt-SNE~\cite{deBodt2018ttSNE}, by reimplementing the respective affinity computation in openTSNE. In tt-SNE the standard high-dimensional Gaussian kernel is replaced with a Student-\(t\) kernel.

For all variants we use perplexity \(\rho=30\) as default. For the multiscale 
models we used perplexities \(\{30, 50, 100, 200\}\). For \texttt{FixedSigmaNN}, 
we set \(\sigma\) to the median distance to the \(\rho\)-th nearest neighbor 
across all points. All variants are evaluated under an identical optimization 
pipeline so that only the affinities differ.

Figure~\ref{fig:global_nh_affinity} (b) shows the results on MNIST. At small \(k\), 
sharpened t-SNE (\(\gamma=1.5\)) gives the highest \(NO@k\), reaching 
approximately \(0.5\) compared to \(0.42\) for standard t-SNE, before dropping 
sharply for larger \(k\). In the mid-local range (\(k \approx 25\)--\(90\)), 
smoothed t-SNE (\(\gamma=0.7\)) outperforms all other methods, including the 
multiscale variants, even though \texttt{MultiscaleMixture} and 
\texttt{Multiscale} incorporate information from multiple perplexity scales. Beyond \(k \approx 90\), \(\gamma=0.0\), 
\texttt{Uniform}, and \texttt{FixedSigmaNN} take over and reach 
\(NO@k \approx 0.40\) at \(k=200\). \(\gamma=0.0\) and 
\texttt{Uniform} indeed produce identical curves. tt-SNE tracks above standard t-SNE from \(k \approx 50\) onwards, consistent with its smoother high-dimensional affinity distribution, and shows 
the flattest curve overall with the least change across \(k\). In contrast, 
\(\gamma=0.0\)/\texttt{Uniform} and \texttt{FixedSigmaNN} show the largest 
variation, rising steeply from low values at small \(k\) to the highest values 
at large \(k\). The curves reveal a consistent trade-off governed by how 
sharply each affinity construction concentrates probability mass on the nearest 
neighbors.

\section{Conclusion\label{sec:conclusions}}
In this work we analyzed the role of affinity sharpness in t-SNE, focusing on how the row-wise distribution of probability mass in the high-dimensional affinity matrix affects neighborhood preservation at different scales. We introduced a row-wise power transform controlled by the parameter \(\gamma\), which smooths or sharpens each conditional probability row. We showed analytically that this transform is equivalent to rescaling the Gaussian bandwidth, but produces point-dependent effective perplexities, making it  different from changing the global perplexity.

Our experiments reveal a clear scale-dependent trade-off. Sharpening the affinity rows improves preservation of the very nearest neighbors, while smoothing improves preservation of broader local neighborhoods. Standard t-SNE, with its fixed-perplexity construction, appears to concentrate probability mass too heavily on the first few neighbors, which limits its ability to preserve broader local structure. Smoothing redistributes this mass and gives more influence to mid-ranked neighbors during optimization, improving mid-local and global structure preservation. The two parameters are complementary: perplexity controls how many neighbors are included, while \(\gamma\) controls how mass is distributed among them. Smoothing has the most effect at lower perplexities, where the affinity distributions are very sharp. We further show that \(\gamma=0.7\) gives the strongest mid-local preservation among the affinity constructions we evaluated, including multiscale methods, while remaining competitive at other scales; whether others could do better there remains open.

These findings suggest that affinity sharpness is a meaningful and previously underexplored axis of t-SNE behavior. The choice of \(\gamma\) provides practitioners with a lightweight way to shift the preservation focus depending on their visualization goal: sharper affinities for nearest-neighbor fidelity, smoother affinities for broader local and global structure.

A limitation of this study is the main focus on a single quality measure (NO@k/\(Q_{NX}(k)\)), although its simplicity gives a clear view of one specific aspect of embedding quality. The improvements are consistent but modest in absolute terms, and establishing when they translate into a practical benefit is itself difficult and beyond our present scope. Secondly, future work could study how to optimize the hyperparameters $\rho, \gamma$ to maximize specific NO@k.

\begin{credits}
\subsubsection{\ackname} This research was funded by the Flemish Government (AI Research Program), the FWO (G073924N), the EU (ERC, VIGILIA, 101142229), and the BOF of Ghent University (BOF20/IBF/117). Views and opinions expressed are however those of the author(s) only and do not necessarily reflect those of the EU or the ERC Executive Agency. Neither the EU nor the granting authority can be held responsible for them. For the purpose of Open Access the authors have applied a CC BY public copyright license to any Author Accepted Manuscript version.
\subsubsection{\discintname}
The authors have no competing interests to declare that are relevant to the
content of this article.
\end{credits}
%
%
%
\bibliographystyle{splncs04}
\bibliography{references}

\end{document}


\title{Appendix: How smoothing the affinity matrix affects neighborhood preservation in t-SNE}

\titlerunning{Appendix: How smoothing affects neighborhood preservation in t-SNE}
\author{Shirin Mohebi \and Guillaume Bied \and Jefrey Lijffijt}

\authorrunning{S. Mohebi et al.}

\institute{Ghent University IDLab, Ghent, Belgium\\
\email{\{Shirin.Mohebi,Guillaume.Bied,Jefrey.Lijffijt\}@ugent.be}}

\maketitle

This appendix provides dataset preprocessing details and full experimental results for all three datasets: MNIST, mouse cortex, and UCI Adult. The main paper reports results on MNIST only due to space constraints. Section~A presents preprocessing details. Section~B presents detailed effective perplexity analysis for all three datasets. Sections~C--G present the full results corresponding to each experiment in the main paper, in the same order: effect of smoothing on local neighborhood preservation, smoothing versus increasing perplexity, sensitivity to \(\gamma\) and \(\rho\), effect of smoothing on global structure preservation, and comparison with alternative affinity constructions.

\section{Dataset preprocessing details}

\ptitle{MNIST} MNIST~\cite{lecun1998mnist} contains images of handwritten digits (\(n=70{,}000\), \(m=784\)). Pixel values are scaled to \([0,1]\) and the data are reduced to 50 principal components before constructing t-SNE affinities, following the preprocessing recommended by Kobak and Berens~\cite{kobak2019}.

\ptitle{UCI Adult} UCI Adult~\cite{kohavi1996adult} is a tabular dataset of US census data (\(n=48{,}842\), \(m=14\)). Numerical variables are standardized and categorical variables are one-hot encoded before constructing t-SNE affinities.

\ptitle{Mouse cortex} The mouse cortex dataset~\cite{Tasic2018} contains single-cell RNA sequencing data from adult mouse cortex, grouped into 133 clusters with a strong hierarchical organization (\(n=23{,}822\)). We use the same preprocessed data as Kobak and Berens~\cite{kobak2019}, which includes sequencing-depth normalization, feature selection, log-transformation, and dimensionality reduction to 50 principal components.

\section{Effective perplexity analysis}

\begin{figure}[ht]
    \centering
    \begin{subfigure}[b]{0.7\linewidth}
        \centering
        \includegraphics[width=\linewidth]{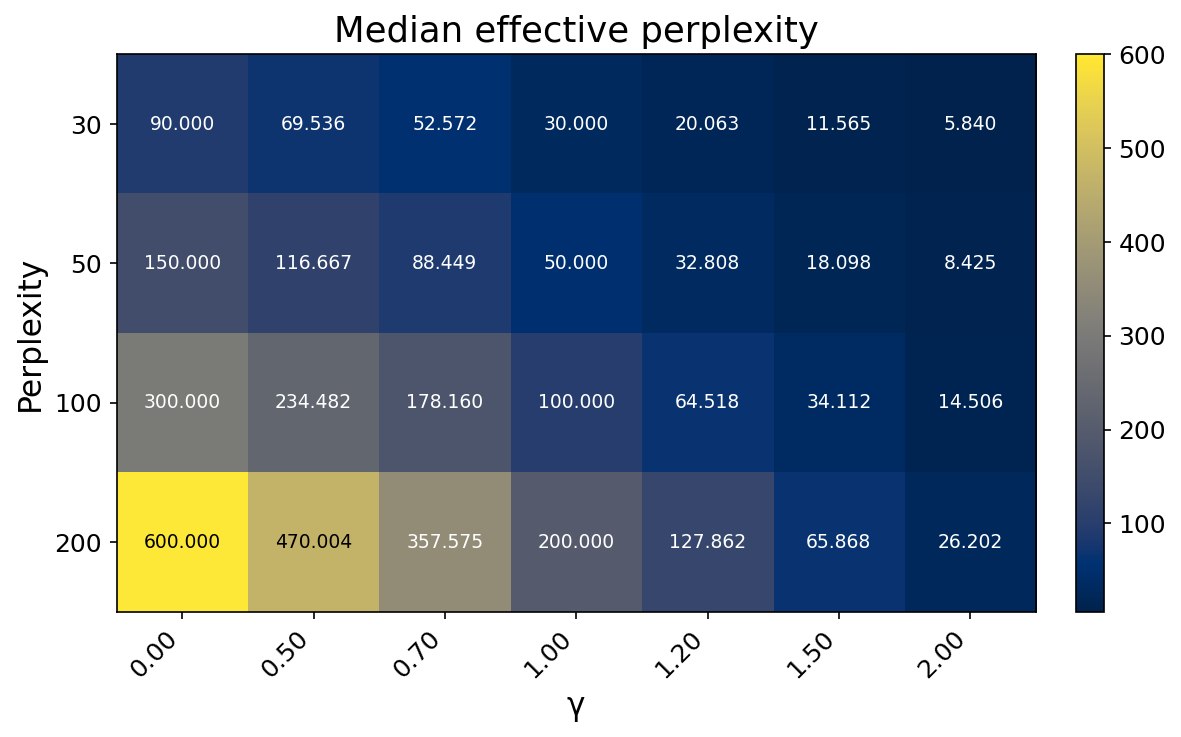}
        \caption{MNIST}
    \end{subfigure}

    \medskip

    \begin{subfigure}[b]{0.7\linewidth}
        \centering
        \includegraphics[width=\linewidth]{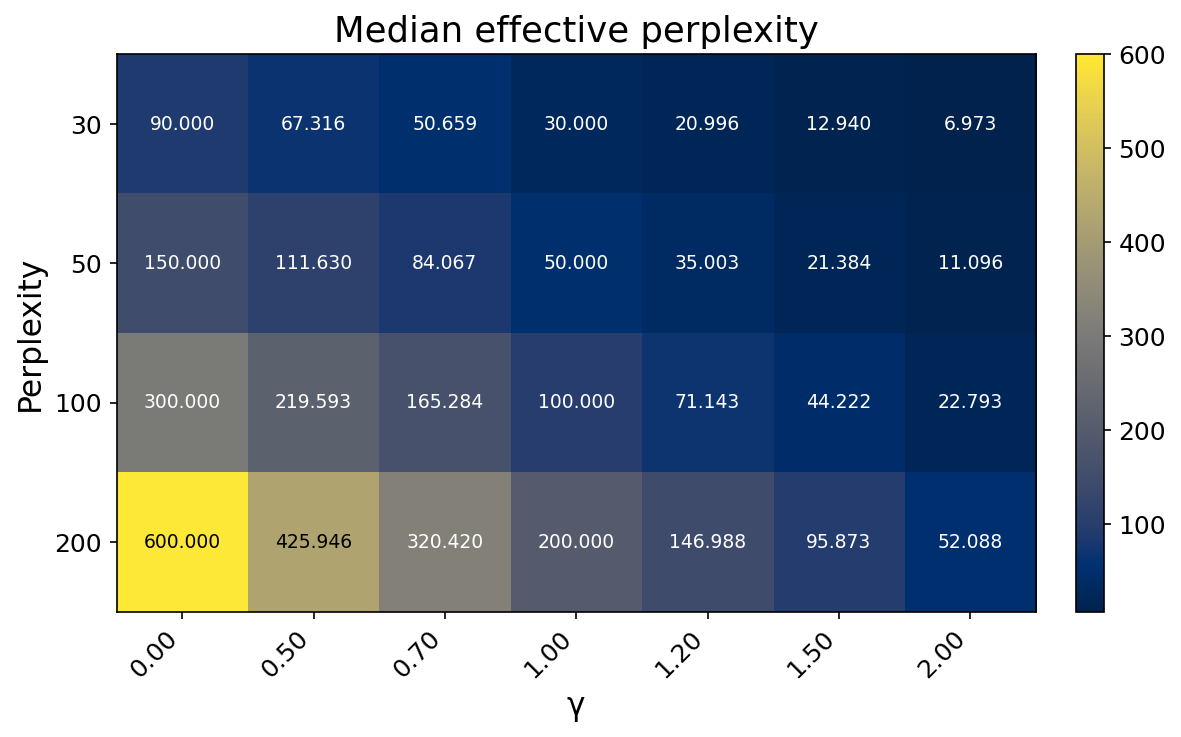}
        \caption{Mouse cortex}
    \end{subfigure}

    \medskip

    \begin{subfigure}[b]{0.7\linewidth}
        \centering
        \includegraphics[width=\linewidth]{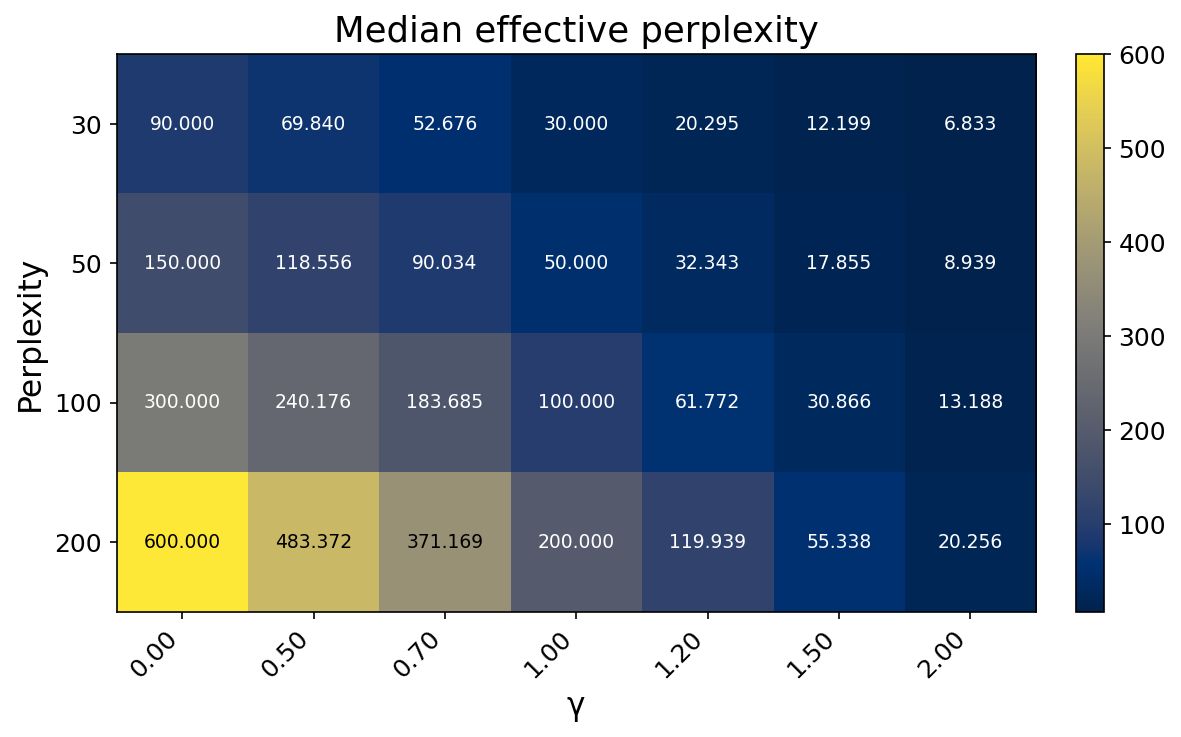}
        \caption{Adult}
    \end{subfigure}
    \caption{Median effective perplexity across \(\rho\) and \(\gamma\) settings on MNIST (a), mouse cortex (b), and Adult (c).}
    \label{fig:app_median_eff_perp}
\end{figure}

Figure~\ref{fig:app_median_eff_perp} shows the median effective perplexity on MNIST, mouse cortex, and Adult across a wider range of \(\gamma\) and \(\rho\) settings. When \(\gamma=1.0\), 
the effective perplexity matches \(\rho\) exactly. Sharpening reduces effective 
perplexity by concentrating the distribution and lowering its entropy, while 
smoothing increases it. The extreme case \(\gamma=0\) produces a uniform 
distribution, giving an effective perplexity of \(3\cdot\rho\), while 
\(\gamma=2.0\) at \(\rho=30\) reduces it to only \(5.840\) on MNIST.

\begin{figure}[p]
    \centering
    \begin{subfigure}{0.48\linewidth}
        \includegraphics[width=\linewidth]{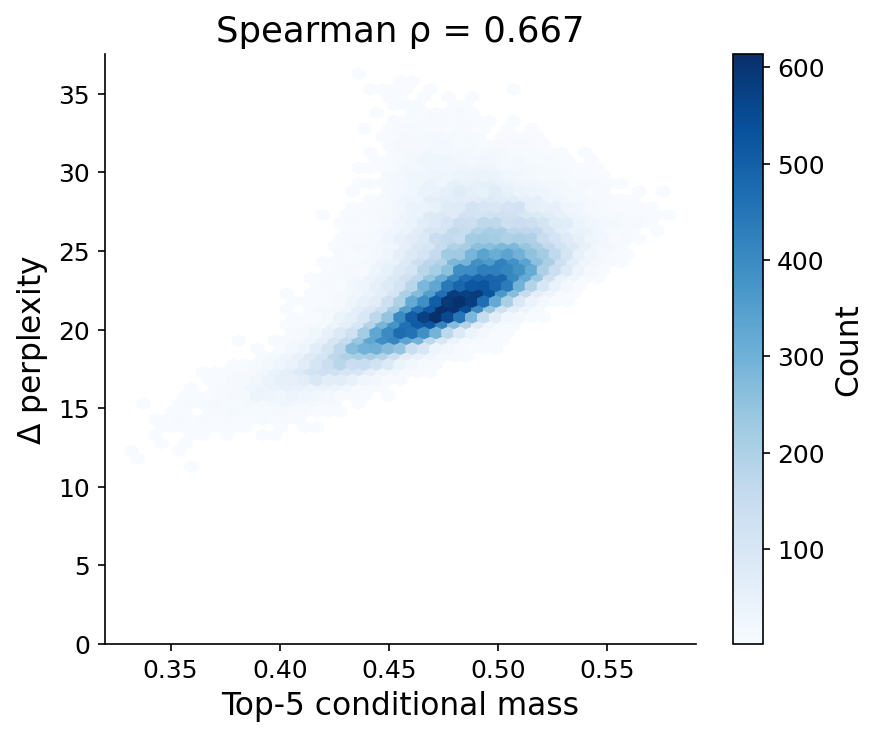}
        \caption{MNIST}
    \end{subfigure}
    \hfill
    \begin{subfigure}{0.48\linewidth}
        \includegraphics[width=\linewidth]{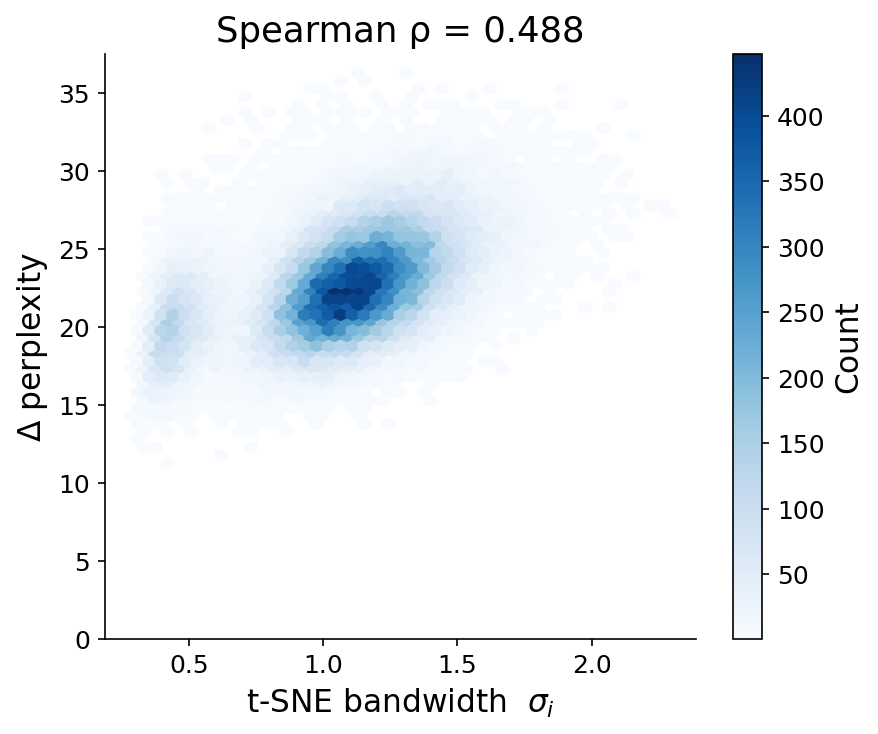}
        \caption{MNIST}
    \end{subfigure}

    \medskip

    \begin{subfigure}{0.48\linewidth}
        \includegraphics[width=\linewidth]{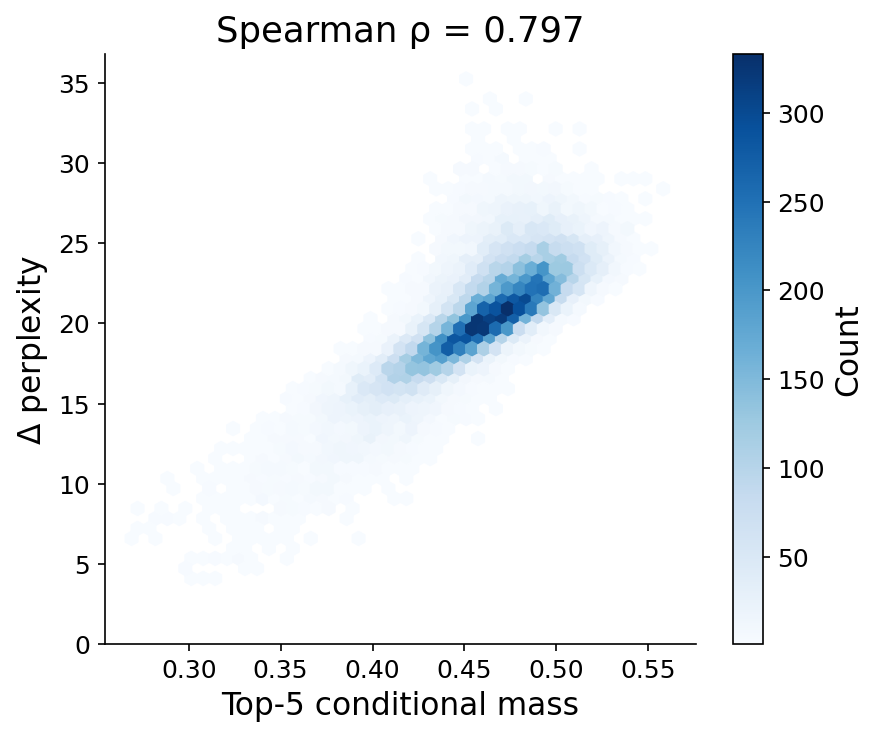}
        \caption{Mouse cortex}
    \end{subfigure}
    \hfill
    \begin{subfigure}{0.48\linewidth}
        \includegraphics[width=\linewidth]{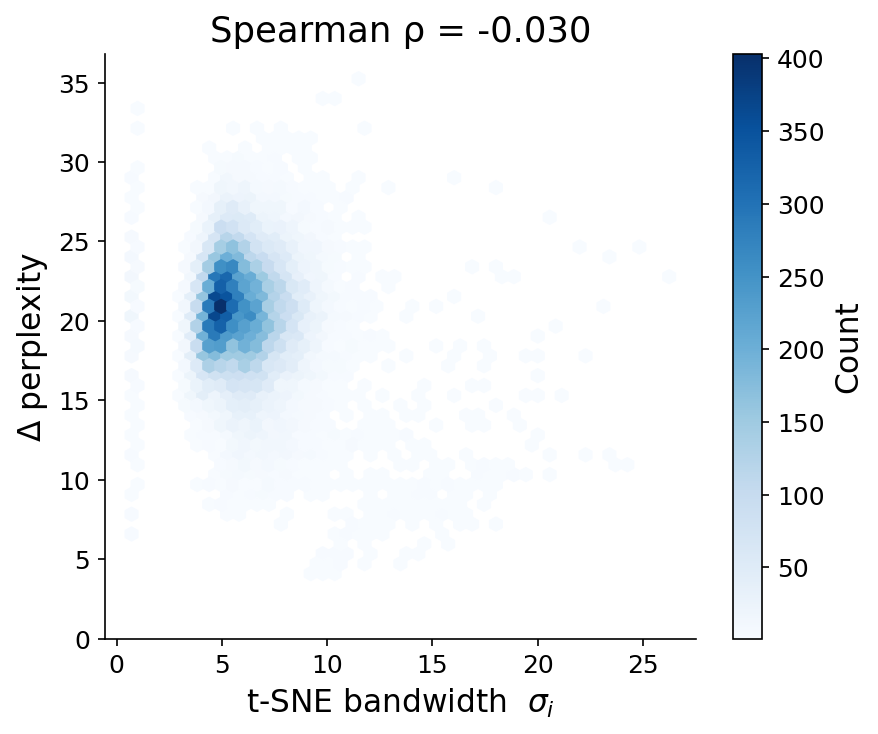}
        \caption{Mouse cortex}
    \end{subfigure}

    \medskip

    \begin{subfigure}{0.48\linewidth}
        \includegraphics[width=\linewidth]{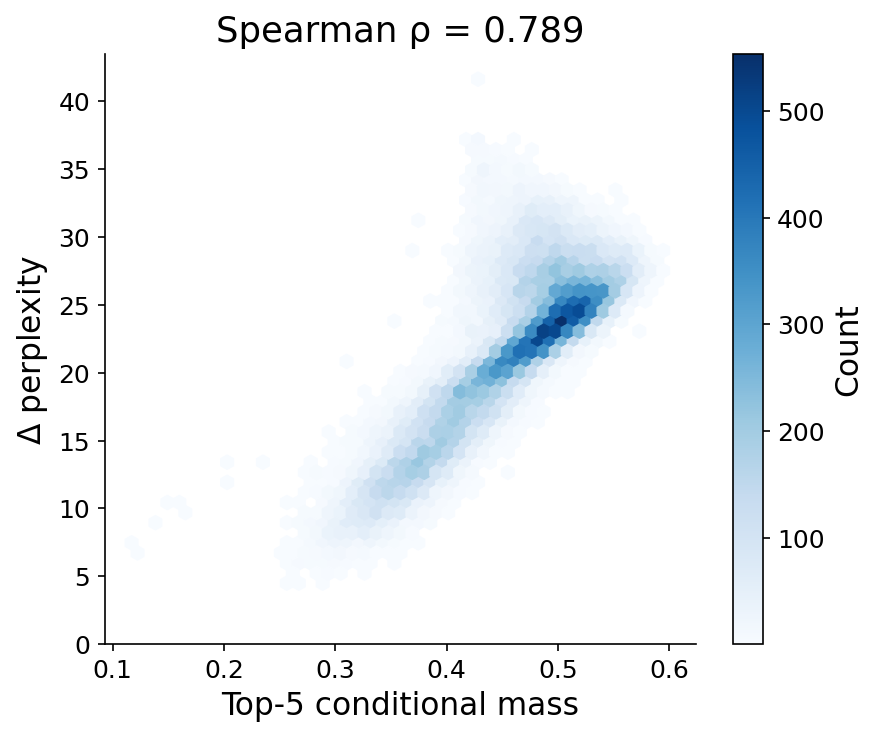}
        \caption{Adult}
    \end{subfigure}
    \hfill
    \begin{subfigure}{0.48\linewidth}
        \includegraphics[width=\linewidth]{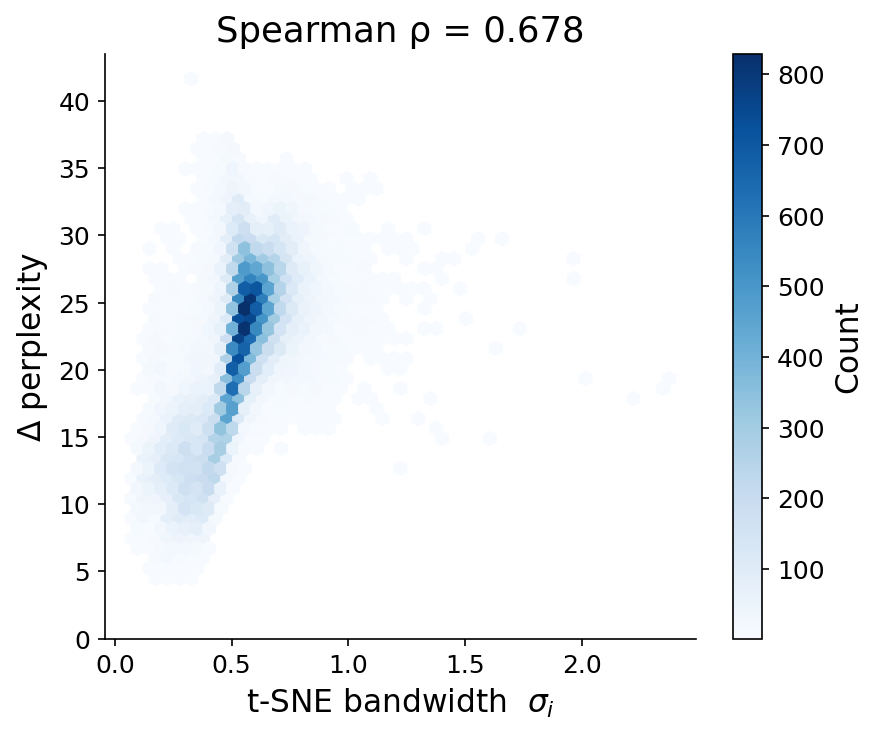}
        \caption{Adult}
    \end{subfigure}
    \caption{Per-point change in effective perplexity after smoothing 
    (\(\gamma=0.7\), \(\rho=30\)). (left) Versus top-5 conditional mass \(M_i(5)\). (right) Versus t-SNE bandwidth \(\sigma_i\) across three datasets.}
    \label{fig:app_perp_pointlevel}
\end{figure}

Figure~\ref{fig:app_perp_pointlevel} shows the per-point change in 
effective perplexity after smoothing (\(\gamma=0.7\), \(\rho=30\)) 
across all three datasets. Since the transform affects each row 
according to its own probability distribution, we investigate which 
point-level properties predict how strongly a point is affected. 
We consider two properties: the top-5 conditional mass \(M_i(5)\), 
which measures how sharply probability mass is concentrated on the 
nearest neighbors, and the standard t-SNE bandwidth \(\sigma_i\), 
assigned before any smoothing is applied, which reflects local 
density around each point.

For the top-5 conditional mass (left column), the pattern is 
consistent across all datasets: points with sharper affinity rows 
receive a larger increase in effective perplexity, with Spearman 
correlations of \(0.667\), \(0.797\), and \(0.789\) for MNIST, 
mouse cortex, and Adult respectively. This is consistent with 
Lemma~1 in the main paper. For the bandwidth \(\sigma_i\) (right 
column), the relationship is weaker and less consistent. For MNIST 
and Adult, sparser points tend to be more affected, with Spearman 
correlations of \(0.488\) and \(0.678\). For mouse cortex the 
relationship is essentially absent (Spearman \(\rho=-0.030\)), 
which may reflect the more complex density structure of that dataset.

\clearpage

\section{Effect of smoothing on local neighborhood preservation}

\begin{figure}[ht]
    \centering
    \begin{subfigure}[b]{0.48\linewidth}
        \centering
        \includegraphics[width=\linewidth]{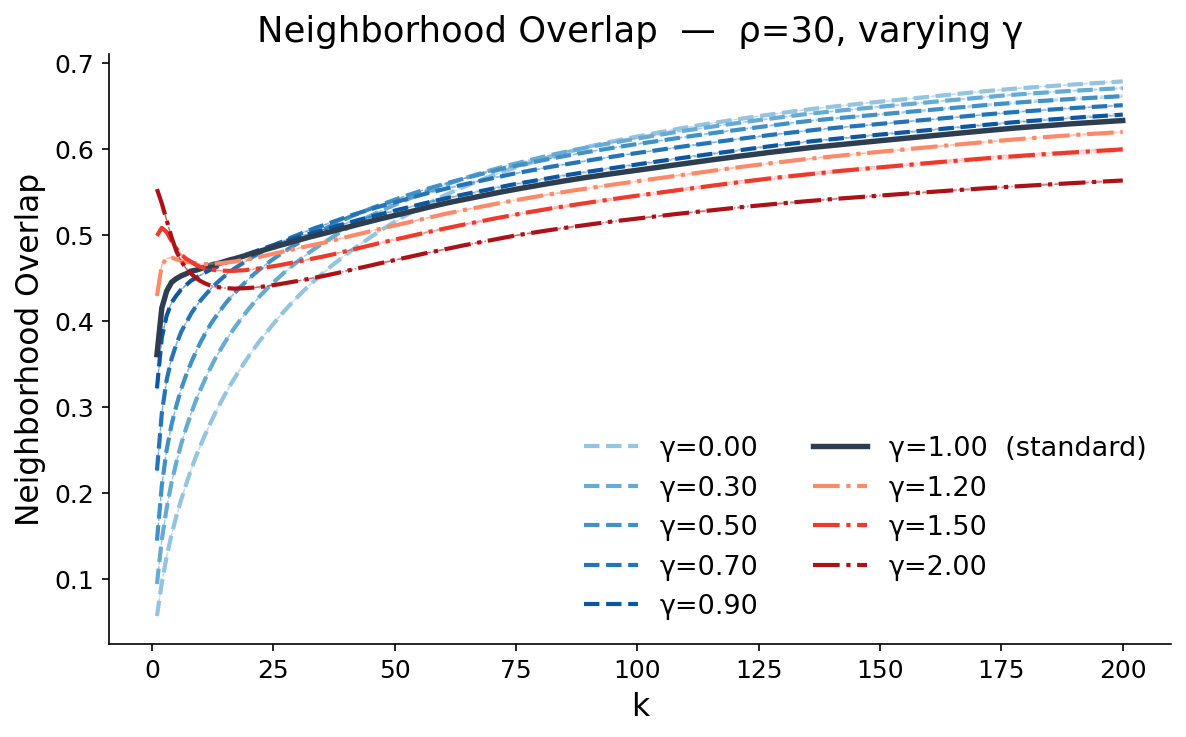}
        \caption{mouse cortex}
    \end{subfigure}
    \hfill
    \begin{subfigure}[b]{0.48\linewidth}
        \centering
        \includegraphics[width=\linewidth]{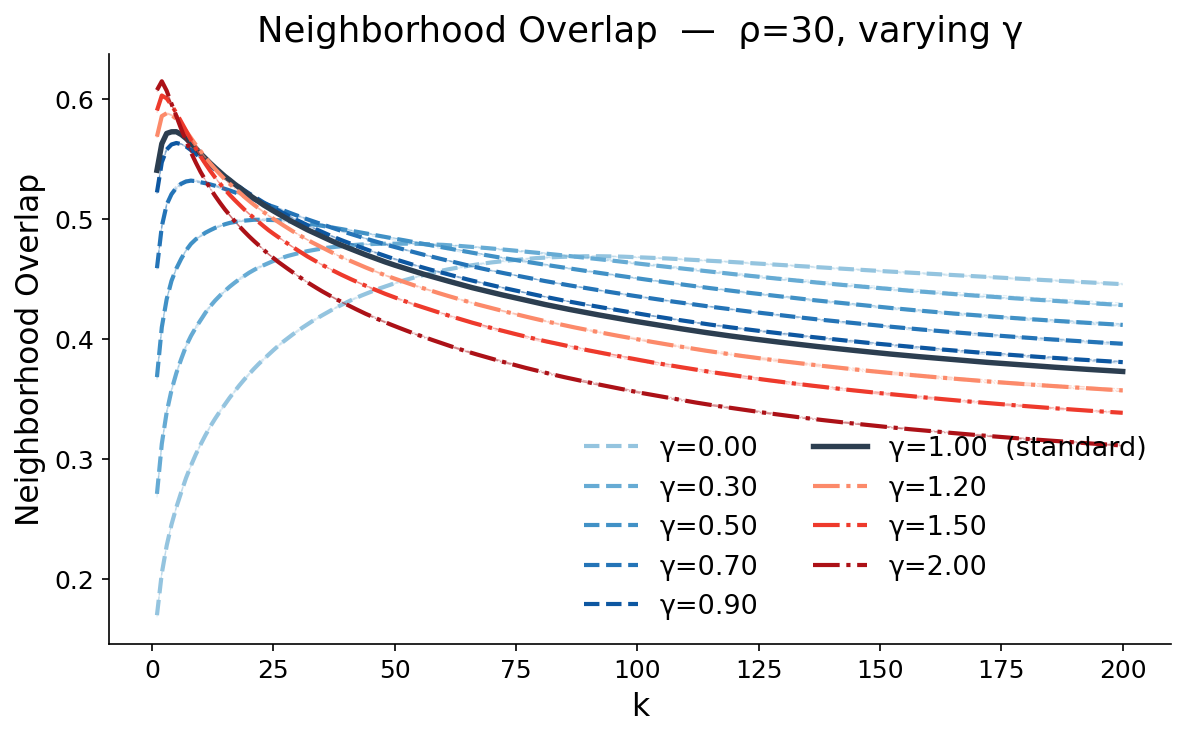}
        \caption{adult}
    \end{subfigure}
    \caption{\(NO@k\) for \(k=1,\ldots,200\) at \(\rho=30\) with 
    varying \(\gamma\), on mouse cortex (a) and Adult (b). The MNIST result is shown in the main paper.}
    \label{fig:app_gamma_sweep}
\end{figure}

Figure~\ref{fig:app_gamma_sweep} shows \(NO@k\) for \(k=1,\ldots,200\) at perplexity \(\rho=30\) for mouse cortex and Adult. Smoothed variants (\(\gamma < 1\)) score better at higher values of \(k\), with the gap growing as \(k\) increases, while sharpened variants (\(\gamma > 1\)) score better at small \(k\). For small values of \(k\), sharpened variants (\(\gamma > 1\)) outperform standard t-SNE. This is because sharpening concentrates more probability mass on the nearest neighbors, giving them stronger attractive force during optimization. For \(k > 10\), this behavior reverses: smoothed variants (\(\gamma < 1\)) consistently outperform both standard t-SNE and sharpened variants, with the gap growing as \(k\) increases. For example in mouse cortex data, \(\gamma=0.5\) reaches approximately \(0.6\) at \(k=100\) compared to \(0.57\) for standard t-SNE and \(0.51\) for \(\gamma=2.0\). We observed the same crossover pattern on all three datasets, confirming that this is not specific to MNIST. This happens because smoothing redistributes probability mass toward mid-ranked neighbors, giving them more influence during optimization rather than letting the nearest neighbors dominate.

\begin{figure}[ht]
    \centering
    
    \includegraphics[width=0.8\linewidth]{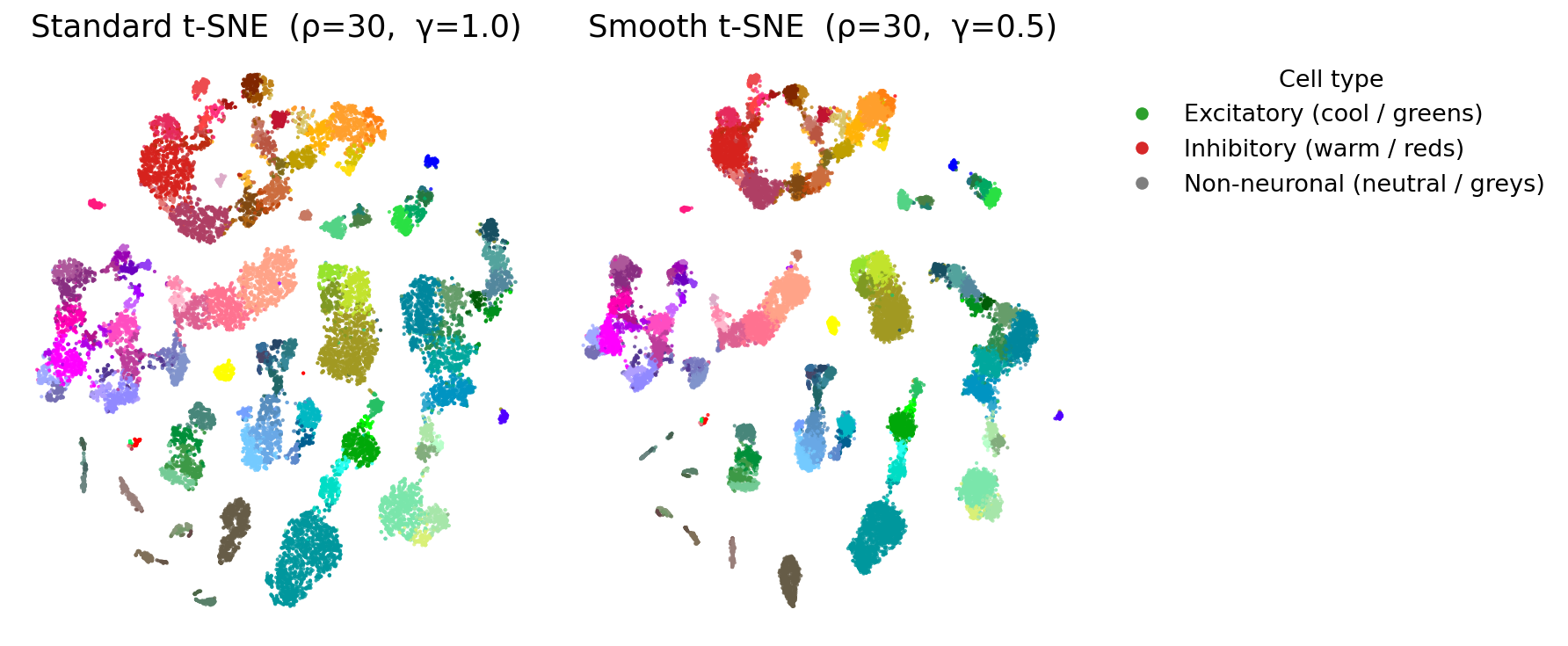}
    \par\smallskip\centering{\small (a) mouse cortex}
    
    \medskip
    \includegraphics[width=0.8\linewidth]{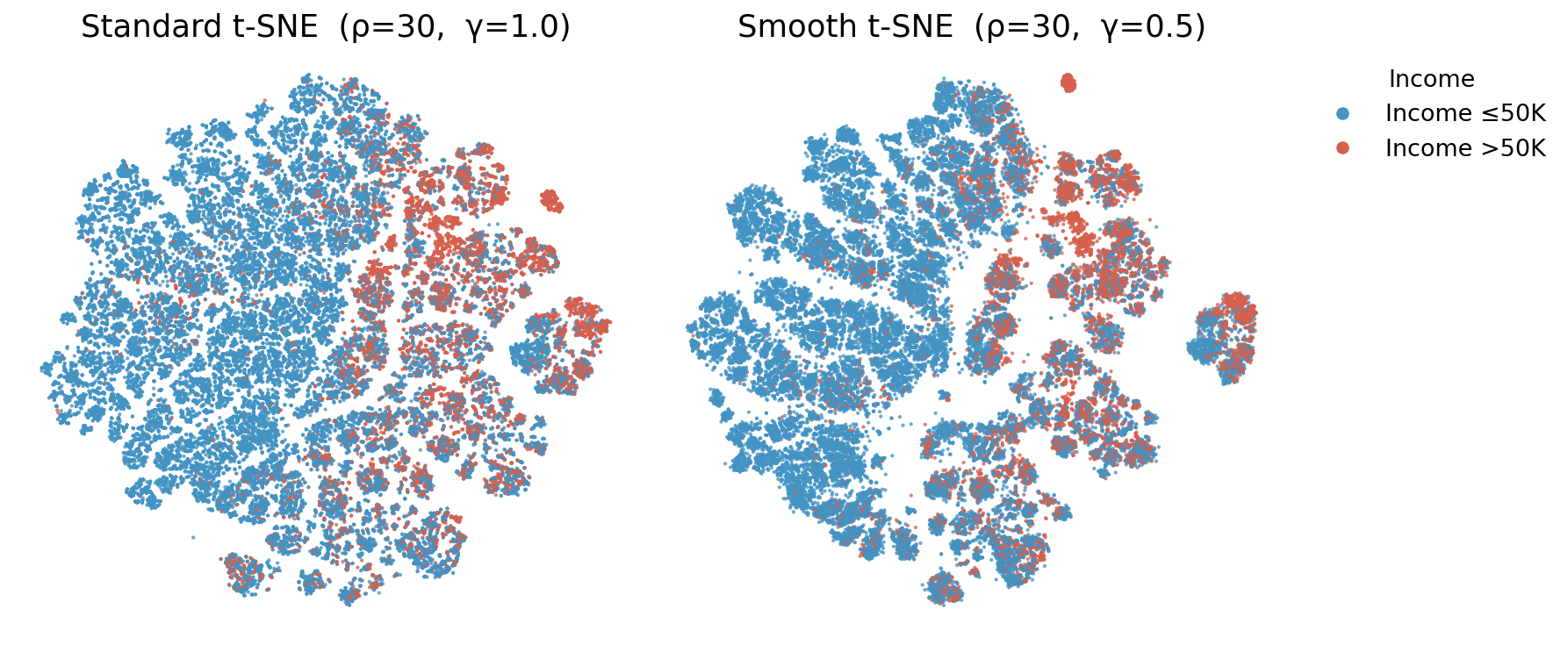}
    \par\smallskip\centering{\small (b) adult}
    
    \medskip
    \caption{Standard t-SNE versus smoothed t-SNE (\(\gamma = 0.5\)) on mouse cortex (a) and Adult (b). The MNIST embedding is shown in the main paper.}
    \label{fig:app_embeddings}
\end{figure}

Figure~\ref{fig:app_embeddings} shows the embeddings for mouse cortex and Adult; the MNIST embedding is shown in the main paper, where the effect of smoothing is most visible, with clusters more separated and digit classes that overlap in standard t-SNE pulled apart. For mouse cortex, the overall hierarchical structure is preserved in both embeddings, with the smooth version showing slightly more separated clusters. For Adult, both embeddings look similar, which is expected: the dataset does not have strong cluster structure, so smoothing has less visible effect on the embedding layout, even though the quantitative neighborhood preservation metrics still show improvement.

\clearpage

\section{Smoothing versus increasing perplexity}
\begin{figure}[h]
    \centering
    \begin{subfigure}[b]{0.48\linewidth}
        \centering
        \includegraphics[width=\linewidth]{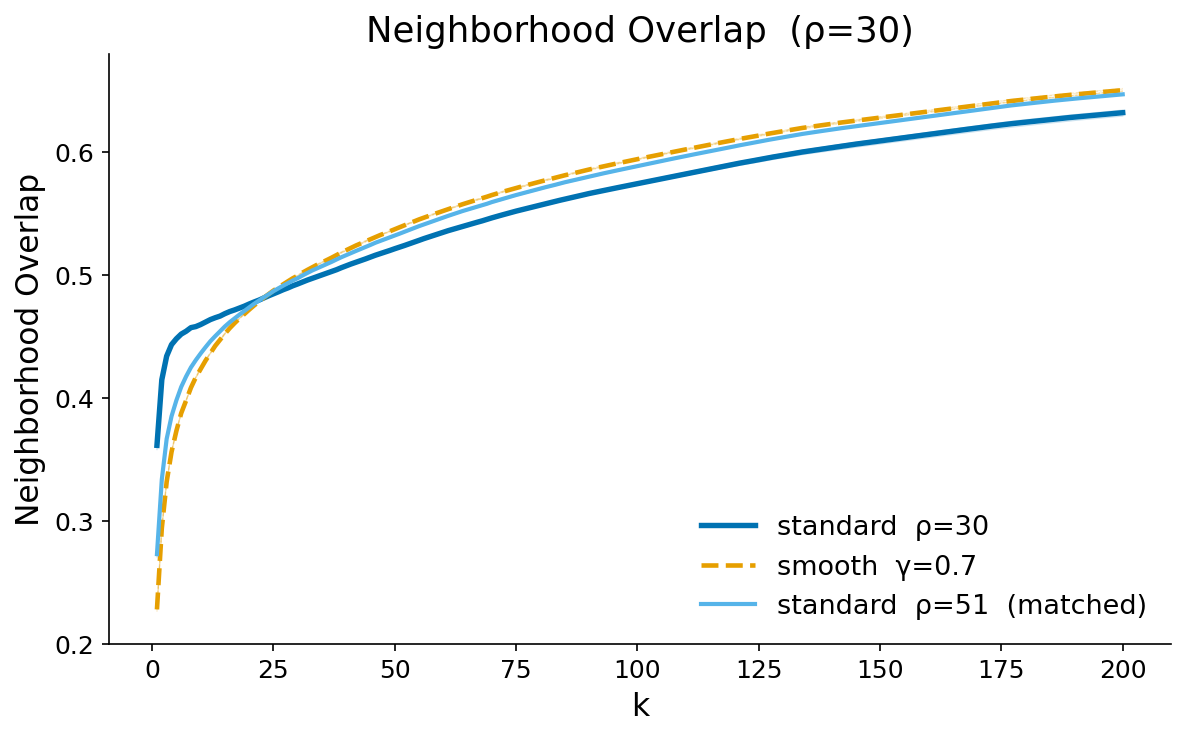}
        \caption{Mouse cortex}
    \end{subfigure}
    \hfill
    \begin{subfigure}[b]{0.48\linewidth}
        \centering
        \includegraphics[width=\linewidth]{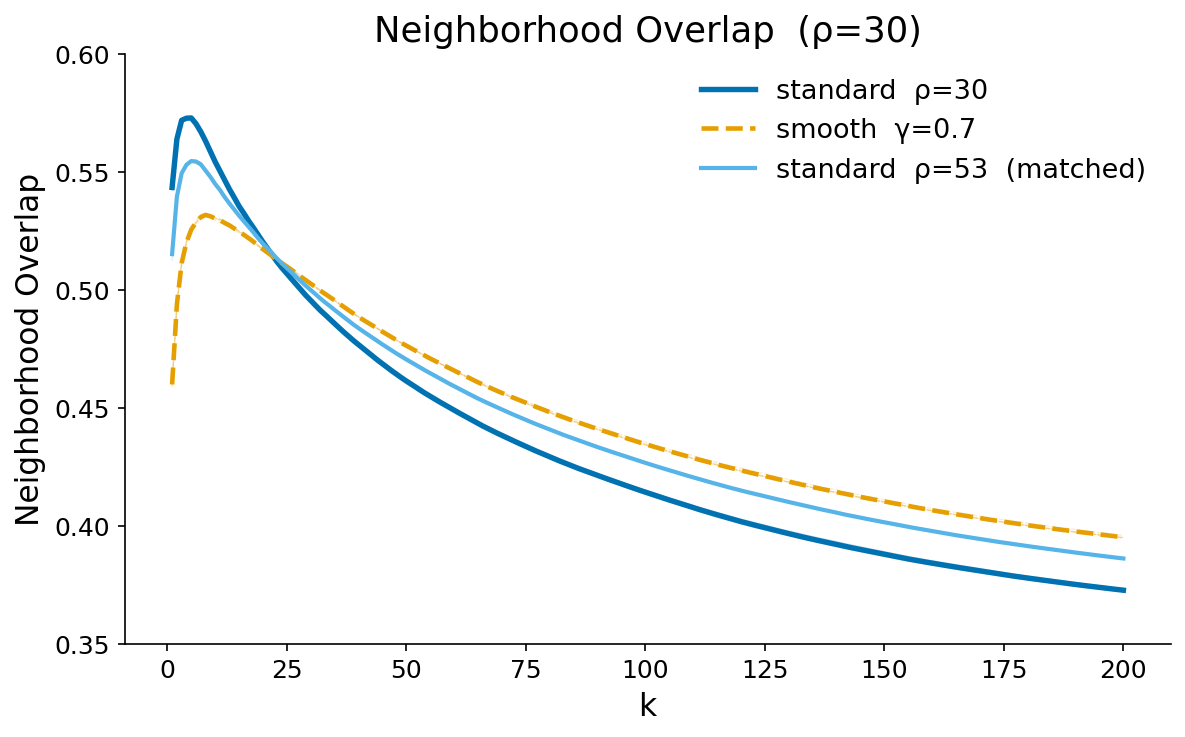}
        \caption{Adult}
    \end{subfigure}
    \caption{\(NO@k\) comparison between standard t-SNE, smoothed 
    t-SNE (\(\gamma=0.7\)), and matched-perplexity standard t-SNE on mouse cortex (a) and Adult (b). The MNIST result is shown in the main paper.}
    \label{fig:app_comparison}
\end{figure}

Figure~\ref{fig:app_comparison} shows the mouse cortex and Adult results;
together with the MNIST result in the main paper, they cover all three datasets.
As shown in the main paper, smoothing widens the effective bandwidth to
\(\tilde\sigma_i = \sigma_i/\sqrt{\gamma}\), which for \(\gamma<1\) raises the
effective perplexity of every conditional row. This raises the question whether
the same neighborhood preservation could be reached simply by using a larger
perplexity. To test this, we compare against standard t-SNE at a perplexity
matched to the median of the per-point effective perplexities of the smoothed run
(\(\rho=30\), \(\gamma=0.7\)), rounded to the nearest integer: \(53\) for MNIST,
\(51\) for mouse cortex and \(53\) for Adult
(Figure~\ref{fig:app_median_eff_perp}). In all cases, smoothed t-SNE outperforms
matched-perplexity standard t-SNE at broader \(k\), despite having the same median
effective perplexity. This confirms that the point-dependent effective
perplexities produced by smoothing lead to different and better mid-local
preservation than simply increasing the global perplexity.

\begin{figure}[h]
    \centering
    \begin{subfigure}[b]{0.48\linewidth}
        \centering
        \includegraphics[width=\linewidth]{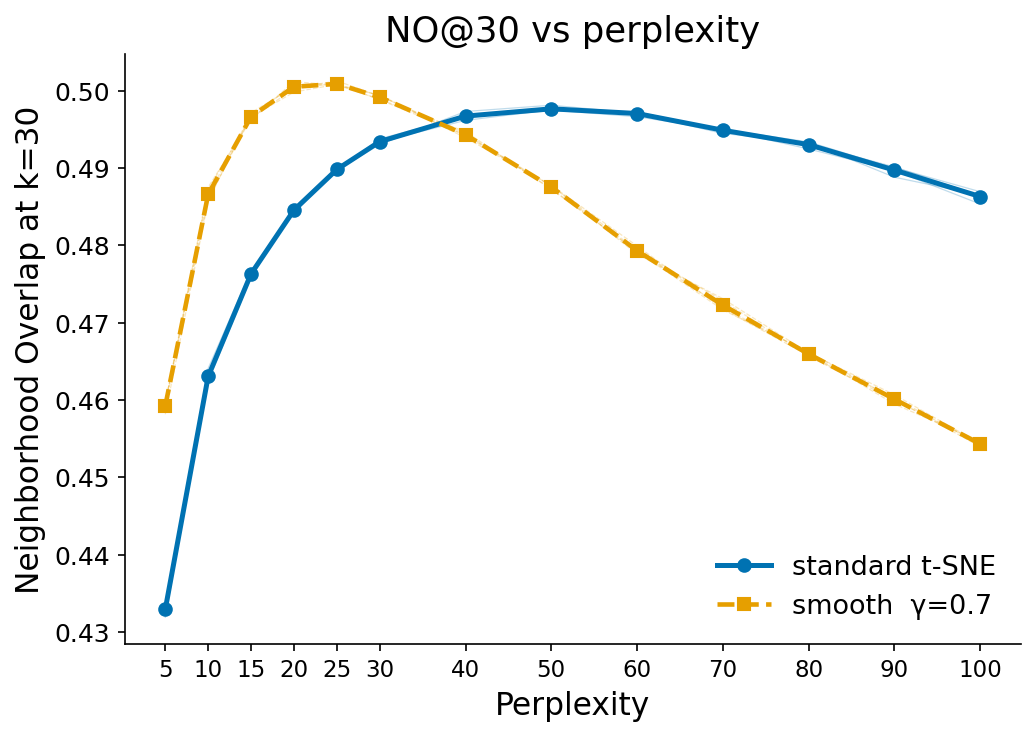}
        \caption{Mouse cortex}
    \end{subfigure}
    \hfill
    \begin{subfigure}[b]{0.48\linewidth}
        \centering
        \includegraphics[width=\linewidth]{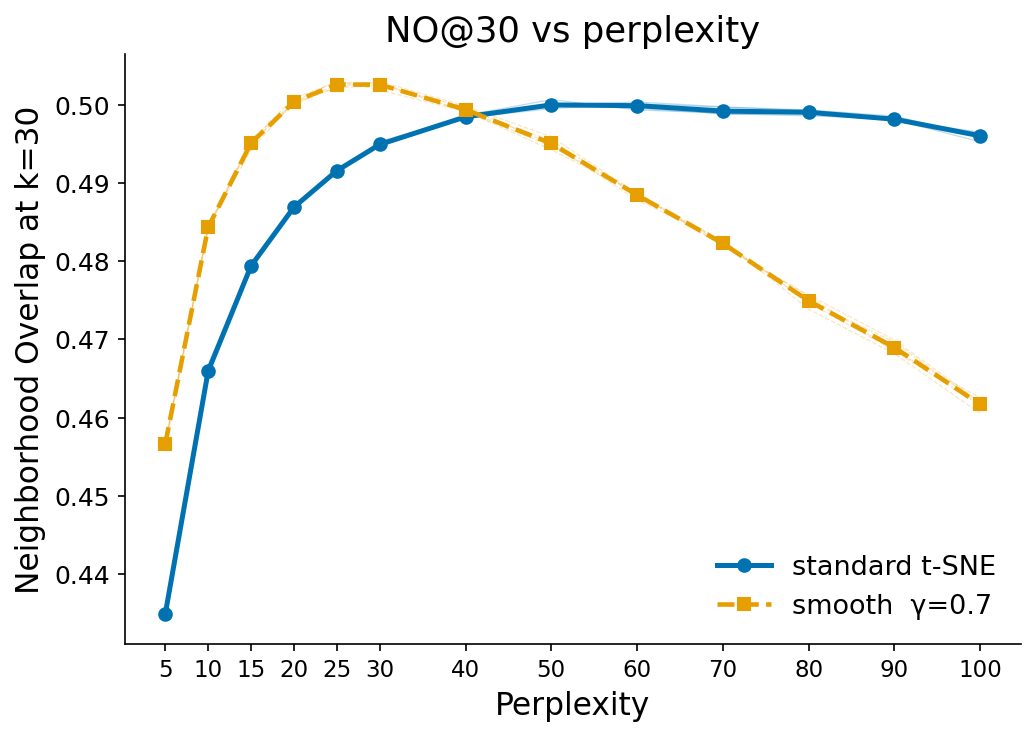}
        \caption{Adult}
    \end{subfigure}
    \caption{\(NO@30\) as a function of perplexity for standard t-SNE 
    and smoothed t-SNE (\(\gamma=0.7\), not tuned) on mouse cortex (a) 
    and Adult (b). Smoothed t-SNE peaks at a lower perplexity than 
    standard t-SNE and attains a higher peak \(NO@30\). The MNIST result 
    is shown in the main paper.}
    \label{fig:app_comparison2}
\end{figure}

Figure~\ref{fig:app_comparison2} shows \(NO@30\) while varying perplexity for standard and smoothed t-SNE (with constant \(\gamma=0.7\), not tuned) on mouse cortex and Adult (the MNIST result is shown in the main paper). The same pattern holds in all cases. Smoothing strongly affects which perplexity is optimal: for standard t-SNE the best perplexity is around \(40\)--\(50\) depending on the dataset, while smoothed t-SNE peaks at a notably lower perplexity of around \(15\)--\(25\). In all three datasets, the highest \(NO@30\) reached by smoothed t-SNE across perplexities exceeds the highest reached by standard t-SNE. However, at higher perplexities smoothed t-SNE degrades more sharply, consistent with the finding that smoothing has adverse effect when the affinity rows are already relatively smooth.

\clearpage

\section{Sensitivity to \(\gamma\) and \(\rho\)}

\begin{figure}[h]
    \centering


    \begin{subfigure}{0.48\linewidth}
        \includegraphics[width=\linewidth]{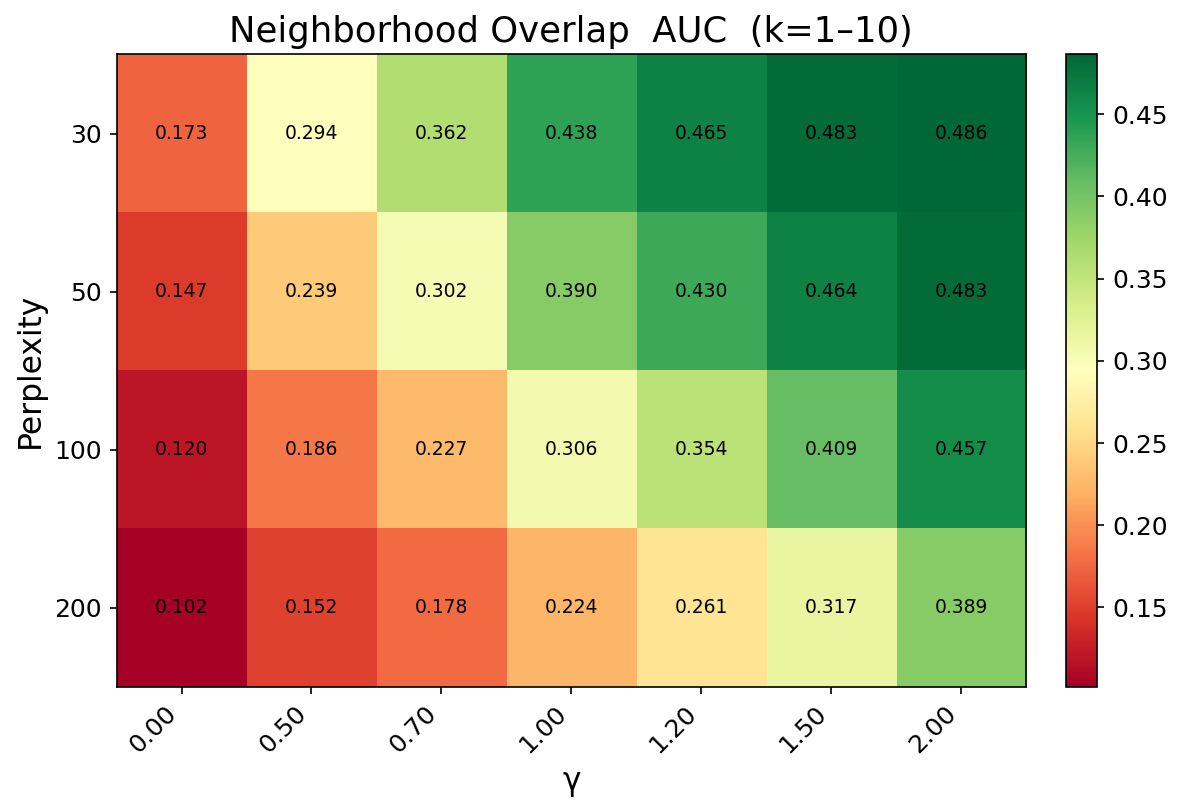}
        \caption{Near-local AUC, Mouse cortex}
    \end{subfigure}
    \hfill
    \begin{subfigure}{0.48\linewidth}
        \includegraphics[width=\linewidth]{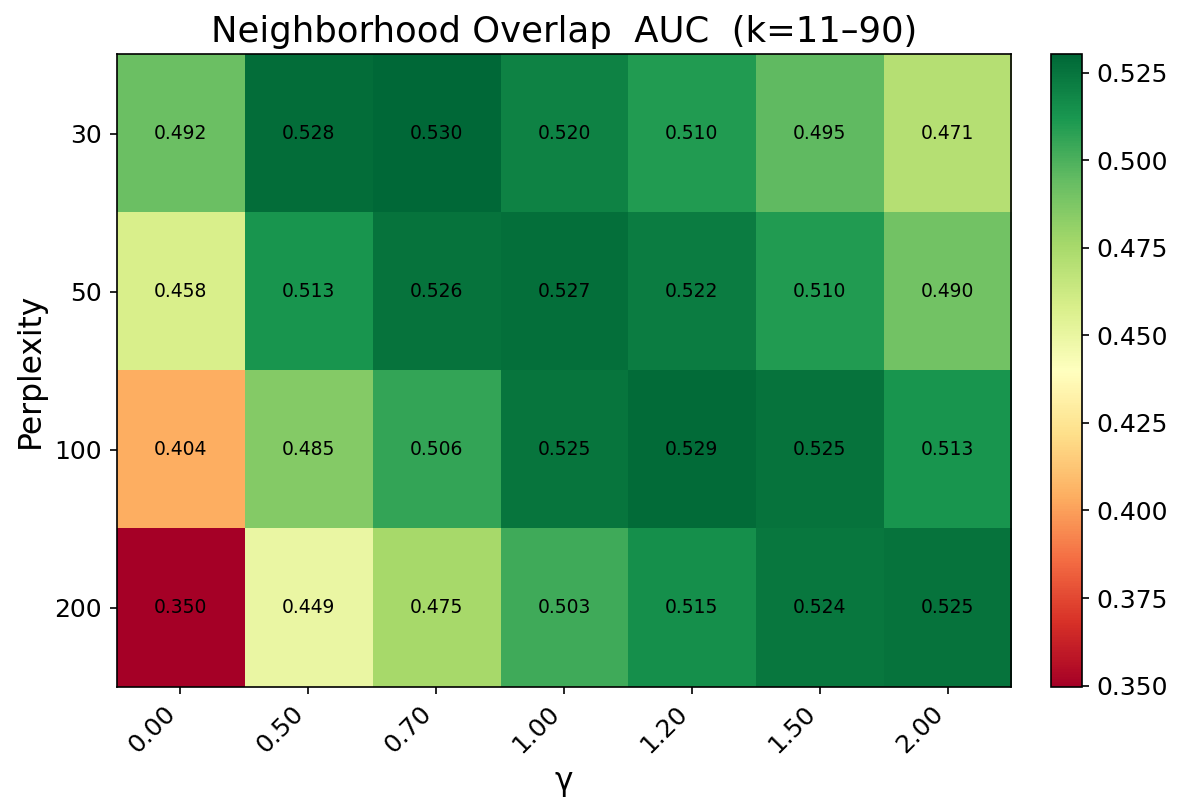}
        \caption{Mid-local AUC, Mouse cortex}
    \end{subfigure}

    \medskip

    \begin{subfigure}{0.48\linewidth}
        \includegraphics[width=\linewidth]{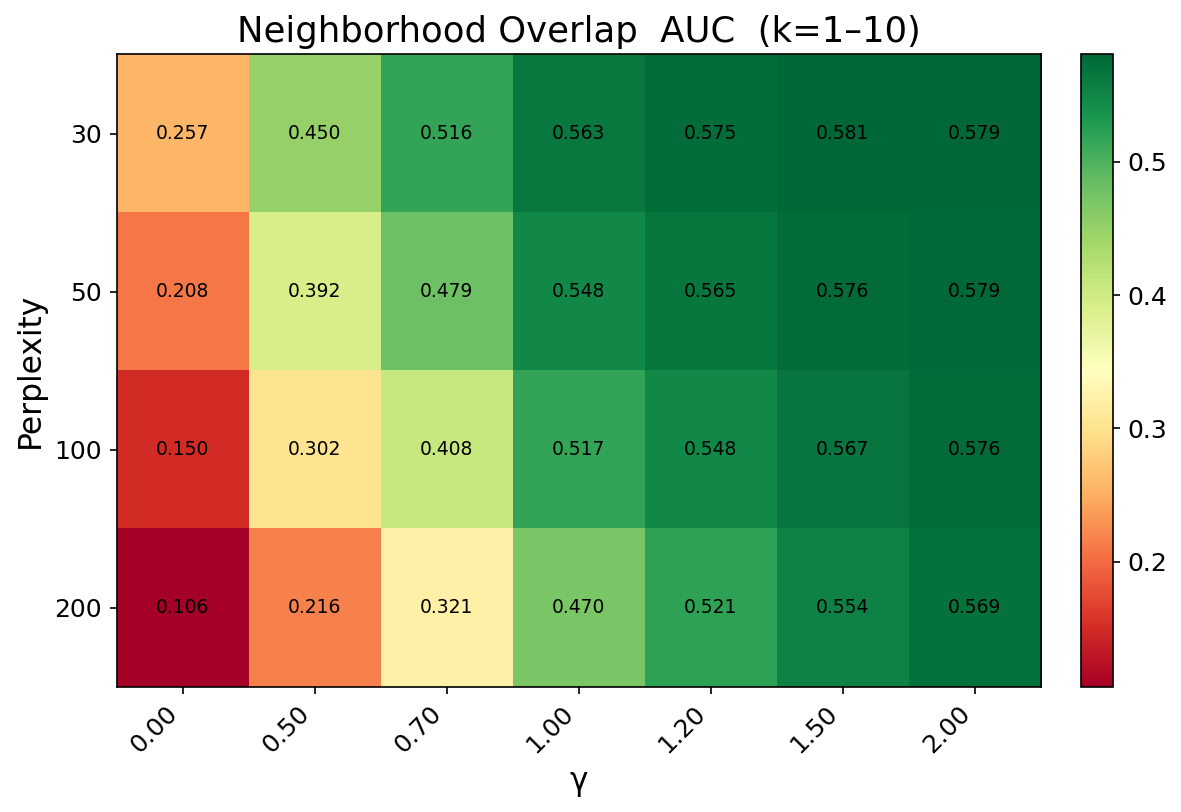}
        \caption{Near-local AUC, Adult}
    \end{subfigure}
    \hfill
    \begin{subfigure}{0.48\linewidth}
        \includegraphics[width=\linewidth]{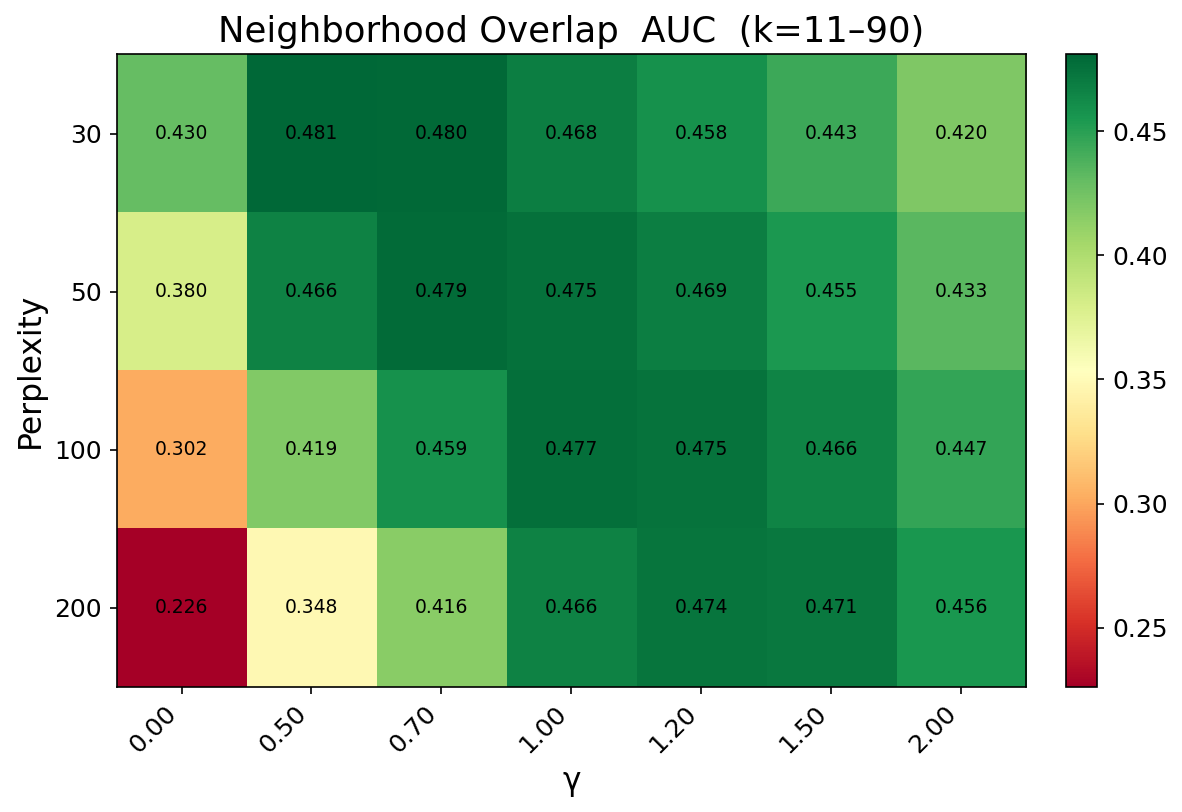}
        \caption{Mid-local AUC, Adult}
    \end{subfigure}
    \caption{Sensitivity of neighborhood preservation to \(\gamma\) 
    and \(\rho\) on mouse cortex (a, b) and Adult (c, d). The MNIST result is shown in the main paper.}
    \label{fig:app_sensitivity}
\end{figure}

Figure~\ref{fig:app_sensitivity} shows the sensitivity results for mouse cortex and Adult; together with the MNIST results in the main paper, they cover all three datasets. For near-local preservation (left column), the pattern is consistent: sharpening improves performance and smoothing hurts it, regardless of dataset or perplexity. This confirms the finding in the main paper holds generally.

For mid-local preservation (right column), the pattern is mostly consistent for MNIST and mouse cortex, where moderate smoothing helps at lower perplexities. Adult shows the same shift, but weaker. Smoothing (\(\gamma=0.7\)) is best only at low perplexity (\(\rho=30\) and \(50\)). At \(\rho=100\) and \(\rho=200\) the best result moves to \(\gamma=1.0\)--\(1.2\), and \(\gamma=0.7\) drops below standard t-SNE.

\clearpage

\section{Effect of smoothing on global structure preservation}

\begin{figure}[h]
    \centering
    \begin{subfigure}[b]{0.48\linewidth}
        \centering
        \includegraphics[width=\linewidth]{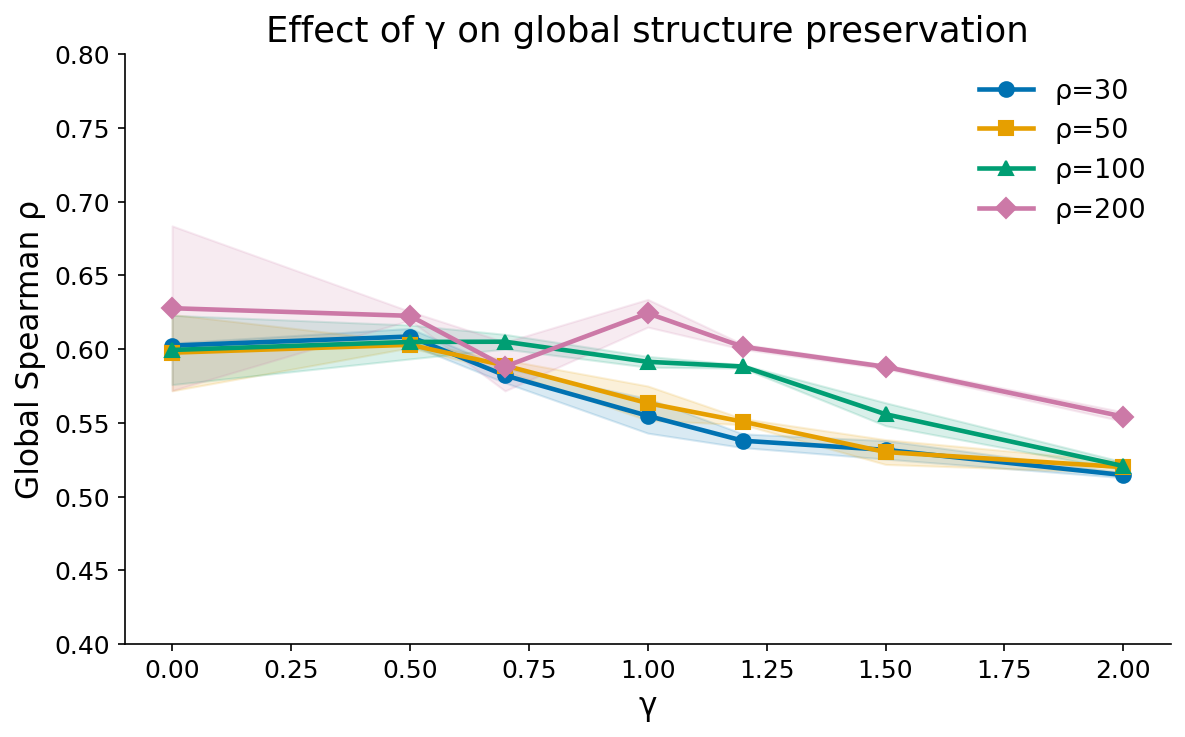}
        \caption{Mouse cortex}
    \end{subfigure}
    \hfill
    \begin{subfigure}[b]{0.48\linewidth}
        \centering
        \includegraphics[width=\linewidth]{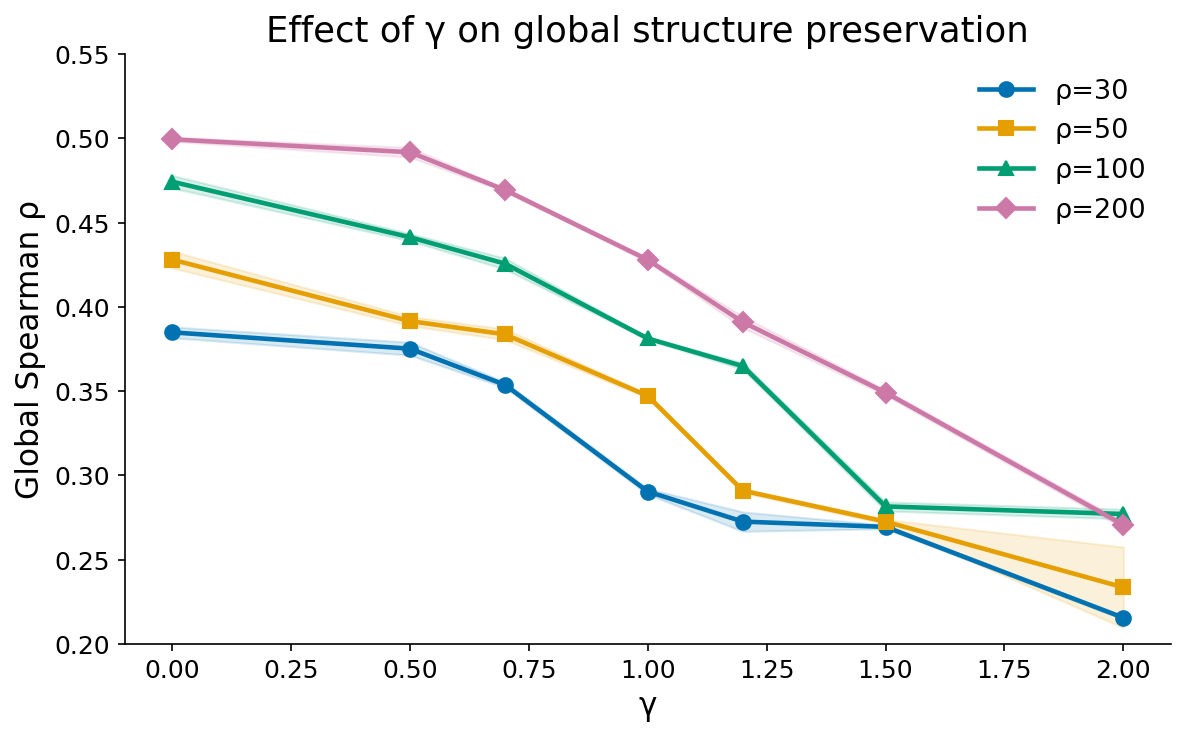}
        \caption{Adult}
    \end{subfigure}
    \caption{Effect of \(\gamma\) on global structure preservation 
    on mouse cortex (a) and Adult (b). The MNIST result is shown in the main paper.}
    \label{fig:app_global}
\end{figure}

Figure~\ref{fig:app_global} shows the effect on global structure preservation for mouse cortex and Adult. The main finding holds across all three datasets (including MNIST in the main paper): sharpening consistently degrades global structure while smoothing generally improves it, most clearly at higher perplexities. Adult shows the strongest absolute values and clearest effect, while MNIST shows the weakest effect, consistent with MNIST having strong local cluster structure that dominates the optimization regardless of \(\gamma\).

\clearpage

\section{Comparison with alternative affinity constructions}
\begin{figure}[h]
    \centering
    \begin{subfigure}[b]{0.48\linewidth}
        \centering
        \includegraphics[width=\linewidth]{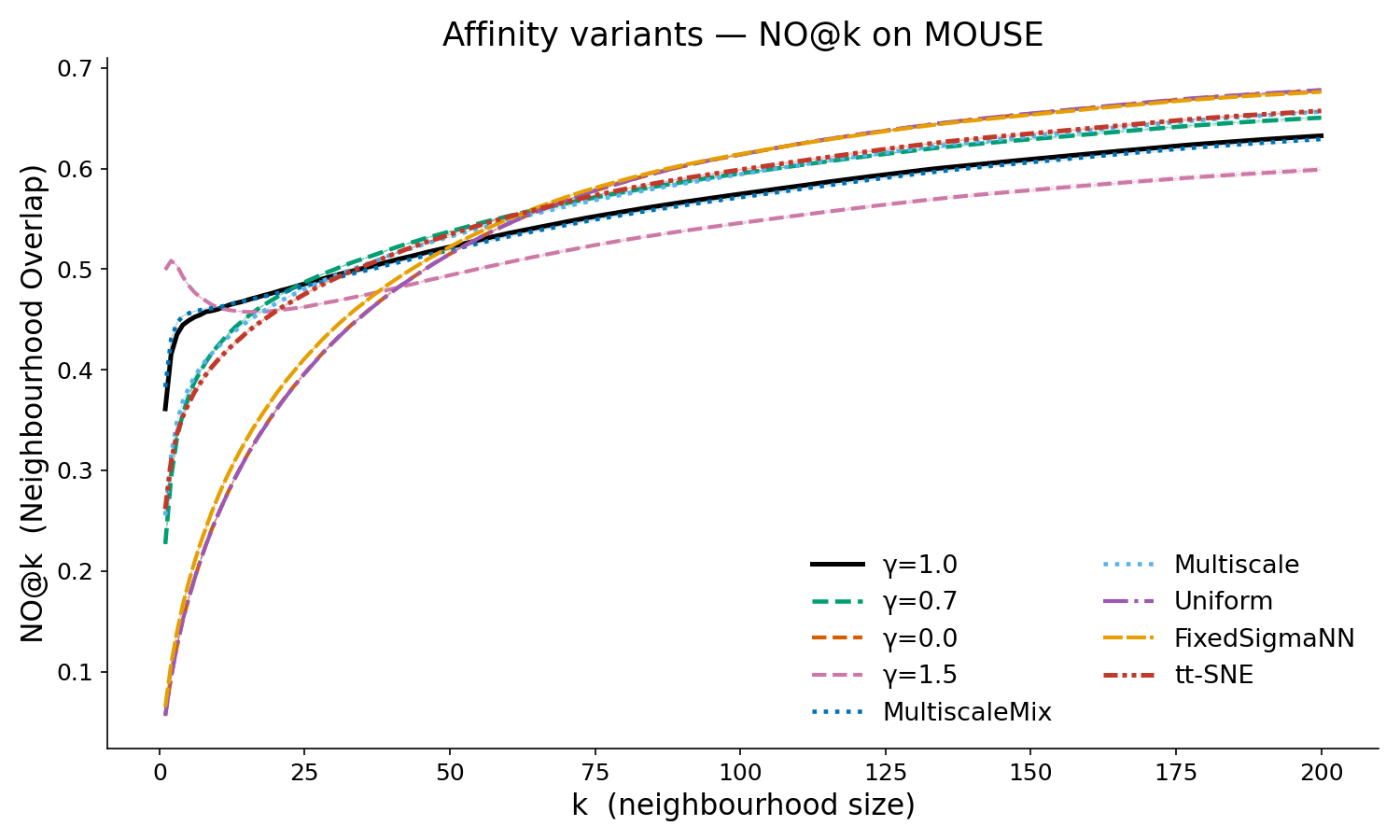}
        \caption{Mouse cortex}
    \end{subfigure}
    \hfill
    \begin{subfigure}[b]{0.48\linewidth}
        \centering
        \includegraphics[width=\linewidth]{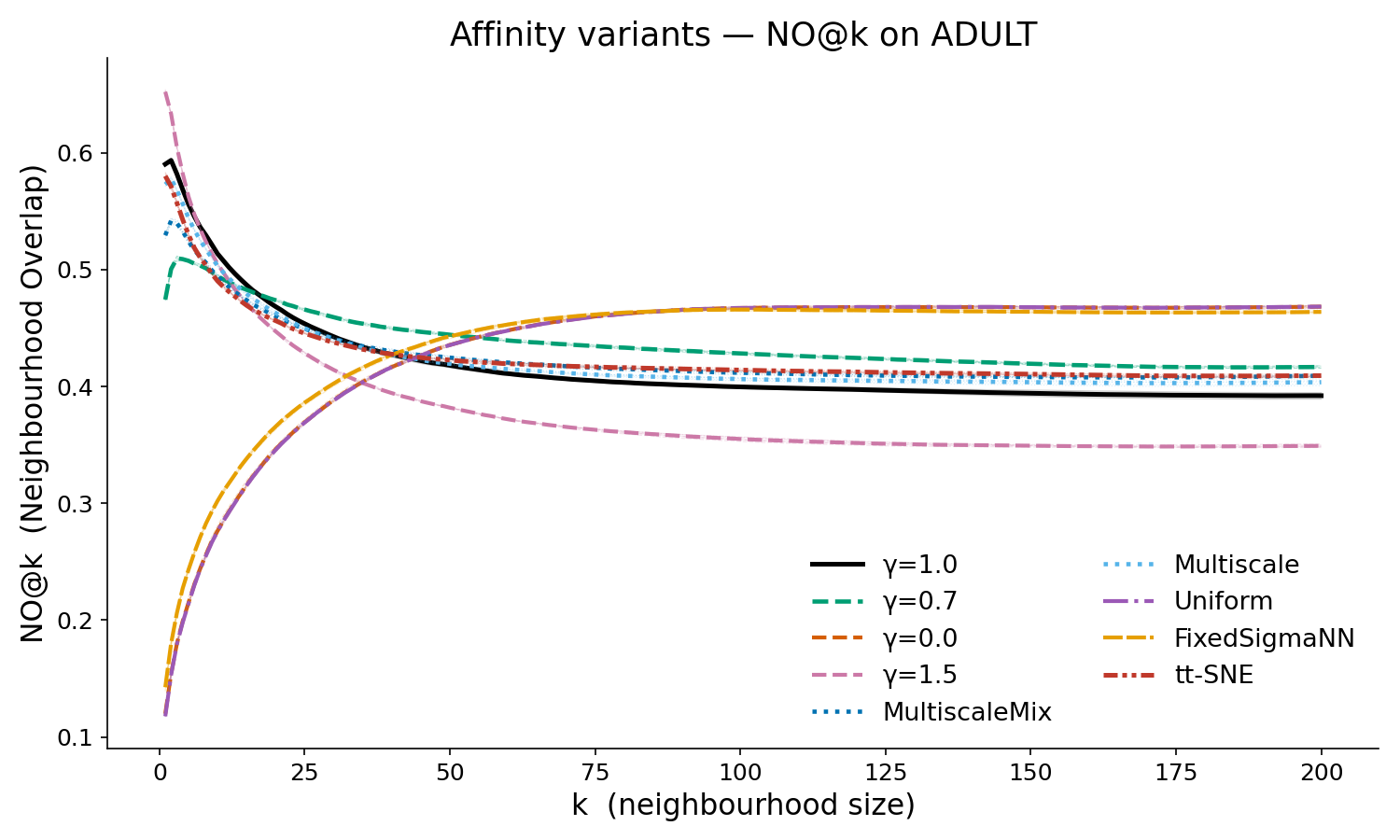}
        \caption{Adult}
    \end{subfigure}
    \caption{{\(NO@k\) for \(k=1,\ldots,200\), comparing \(\gamma\) variants against alternative affinity constructions on mouse cortex (a) and Adult (b). The overall pattern from MNIST holds, with \(\gamma=0.7\) outperforming other methods in the mid-local range. On Adult, \(\gamma=0.7\) maintains its advantage more broadly, while on mouse cortex the multiscale methods are more competitive at larger \(k\).}}
    \label{fig:nh_comp_affinities}
\end{figure}

Figure~\ref{fig:nh_comp_affinities} shows the affinity variant comparison 
on mouse cortex and Adult. The overall pattern from MNIST holds across both 
datasets: \(\gamma=1.5\) leads at small \(k\) and \(\gamma=0.7\) outperforms 
standard t-SNE in the mid-local range. On mouse cortex, the multiscale methods 
are more competitive than on MNIST, tracking closely with \(\gamma=0.7\) in 
the mid-to-large range, while \(\gamma=0.0\), \texttt{Uniform}, and 
\texttt{FixedSigmaNN} again dominate at large \(k\). Adult shows a distinct 
pattern: \(\gamma=0.7\) maintains its advantage across a wider range of \(k\), 
remaining competitive with \(\gamma=0.0\) and \texttt{FixedSigmaNN} even at 
large \(k\), while the multiscale methods offer little improvement over 
standard t-SNE. Across all datasets, \(\gamma=1.5\) ends with the lowest 
\(NO@k\) at large \(k\), with the drop most pronounced on Adult.

\bibliographystyle{splncs04}
\bibliography{references}